\documentclass[10pt,twocolumn,letterpaper]{article}

\usepackage[pagenumbers]{cvpr} 

\usepackage{cuted} 

\usepackage{currfile} 

\usepackage{caption} 

\usepackage[table]{xcolor}
\usepackage{multirow}
\usepackage{hhline}
\usepackage{booktabs}   
\usepackage{multirow}  
\usepackage{pifont}   
\usepackage{xcolor}     
\usepackage{array}     
\usepackage{booktabs}

\newcommand{\method}{PointX\xspace}
\newcommand{\methodbf}{\textbf{PointX}\xspace}

\definecolor{bestblue}{RGB}{186, 216, 255}     
\definecolor{secondblue}{RGB}{214, 232, 255}   
\definecolor{thirdblue}{RGB}{238, 246, 255}    
\newcommand{\best}[1]{\cellcolor{bestblue}{#1}}
\newcommand{\secondbest}[1]{\cellcolor{secondblue}{#1}}
\newcommand{\thirdbest}[1]{\cellcolor{thirdblue}{#1}}

\usepackage[table]{xcolor}
\newcolumntype{G}{>{\columncolor{blue!5}}c}

\definecolor{constructgreen}{RGB}{203, 234, 208}

\definecolor{cvprblue}{rgb}{0.21,0.49,0.74}
\usepackage[pagebackref,breaklinks,colorlinks,allcolors=cvprblue]{hyperref}
\usepackage{soul}

\title{PointX: Multi-View Consistent Intrinsics by Querying Surface Points}

\author{%
    Miguel\,Farinha$^{1,2,*}$\,
    Sam\,Sartor$^{1}$\,
    Iliyan\,Georgiev$^{1}$\,
    Giuseppe\,Vecchio$^{1}$\,
    Ronald\,Clark$^{2}$\,
    Michael\,Fischer$^{1}$
    \\[0.9em]
    $^{1}$Adobe Research
    \qquad
    $^{2\,}$University of Oxford
}

\begin{document}

\twocolumn[{
    \maketitle
    \centering
    \includegraphics{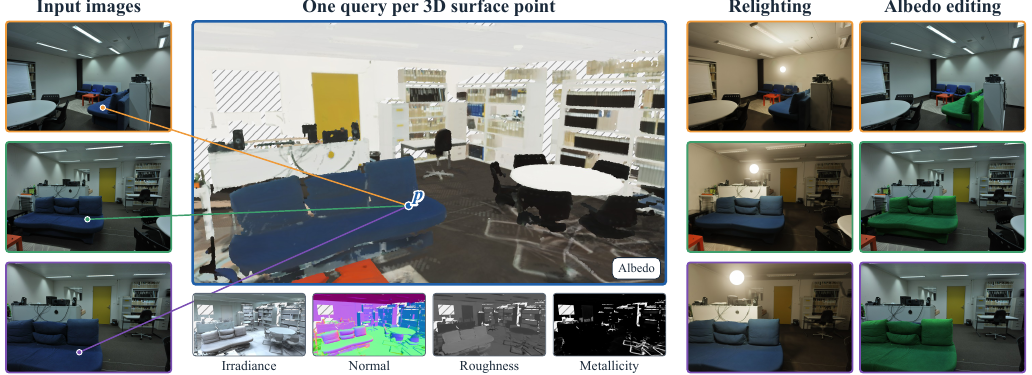}
    \vspace{-4mm}
    \captionsetup{type=figure}
    \captionof{figure}{%
        \methodbf is a feed-forward model for multi-view inverse rendering. From a set of images, it predicts the albedo, irradiance, roughness, metallicity and normal of any queried 3D surface point, here the points of a laser scan~\cite{schops2017multi} registered to the images. Each surface point is decoded once, so the prediction is consistent across views by construction, enabling view-consistent relighting and editing.
    }
    \label{fig:teaser}
    \vspace{6mm}
}]

{\renewcommand{\thefootnote}{\fnsymbol{footnote}}\footnotetext[1]{Work done during an internship at Adobe Research.}}

\begin{abstract}
The appearance of a scene is determined by the albedo, roughness, metallicity and normal of every point on its surface, and by the light that reaches it, its irradiance.
Recovering these intrinsics from images is ambiguous, and methods that predict them per pixel give the same surface point a different decomposition in every view.
We introduce \method, a simple feed-forward model that predicts intrinsics per 3D surface point instead.
A multi-view transformer encodes a set of unposed images, and for any queried surface point, \method pools the point's tokens across the views in which it appears and decodes the point's intrinsics once.
Each point of a 3D reconstruction of the scene, such as a point cloud or a mesh, thus receives a single prediction, which is exactly consistent across views by construction and enables view-consistent relighting and editing.
The same model also predicts per-view maps, from one image or many, by querying the point seen at every pixel.
These maps agree across views up to the precision of the estimated geometry and correspondences.
\method is significantly more consistent than existing methods, while remaining competitive in accuracy from a single image and improving as views are added.
\par\noindent Project page: \small\url{https://pointx-paper.github.io/}
\end{abstract}
\section{Introduction}
\label{sec:intro}

Inverse rendering decomposes an image into the inputs of a renderer: the albedo, normal, roughness and metallicity of every surface point, and the light that reaches it, its irradiance.
With these intrinsic properties~\cite{barrow1978recovering}, a captured scene can be relit, its materials edited, or new objects inserted into it.
However, the decomposition is ambiguous, since the same pixel color has several valid explanations: a wall that appears red may be white under red light or red under white light. While recent methods~\cite{li2020inverse, kocsis2024intrinsic, zeng2024rgb, sartor2025teamwork, careaga2024colorful, ke2025marigold, pavlovic2026marigoldv2, gu2026intrinsync, zeng2026rgbxnext} achieve accurate single-view predictions, a scene is usually captured in several images, or a video.
The same surface point appears in several of them, yet per-image methods resolve the ambiguity in each image independently, so the point receives a different decomposition in every view.
Applications using these intrinsics across multiple views then inherit this disagreement: a relit scene flickers from view to view, an edit made in one image does not carry over to the next, and materials texturing a reconstruction show seams.

One line of work addresses this inconsistency by attaching the intrinsics to a 3D representation of the scene, fitted per scene~\cite{zhang2021nerfactor, jin2023tensoir, zhang2023neilf++, gao2024relightable, liang2024gs, niu2026sgs} or predicted by a network~\cite{li2025lirm, karpikova2026invsplat, schmid2026genrecon, zhang2026worldrecon}, and rendering them into every view, so that all views show the same underlying intrinsics.
However, every view then inherits the errors of this representation, and per-scene fits take minutes to hours.
A second line of work drops the 3D representation.
Multi-view networks~\cite{wu2026mvinverse, li2025idarb, du2026mvid} compute the features of each view from all images through cross-view attention, and video models do the same across frames~\cite{liang2025diffusionrenderer, vrgbx2025, sun2026unid, zeng2026rgbxnext}.
Their predictions are more consistent than those of single-view networks, but each view is still decoded independently, so corresponding pixels of different views can still disagree.
A third line of work keeps the per-view predictions and reconciles them afterwards, through a multi-view consensus at test time~\cite{dirik2026geoid} or a per-scene optimization~\cite{kocsis2026intrinsic}, which has to resolve the disagreement of separately predicted views.
Thus far, consistency has either required an explicit 3D representation, been encouraged through attention and training, or been enforced after the predictions are made.

Our key insight is to predict intrinsic properties per surface point rather than per pixel.
We introduce \methodbf, a feed-forward model that takes one or several unposed images of a static scene and predicts the intrinsics of a query surface point using evidence from all views in which it appears.
A multi-view transformer, initialized from a feed-forward reconstruction model~\cite{wang2026vggtomega}, first computes a feature map for each image.
Given a query point in 3D or pixel space, we then find its correspondences across the input images and sample the feature maps at these locations.
A 3D point is projected into each image with cameras and depth, estimated by a frozen reconstruction model~\cite{wang2026vggtomega} or taken from any other source, and a pixel can also be matched across the images with a dense matcher~\cite{nordstrom2026romaomega}.
We pool these features into a single representation of the surface point, weighting each view by the confidence that the point is visible in it, and small MLP heads decode the point's intrinsic properties from it only once.
The resulting prediction belongs to the point rather than to a view, making it consistent by construction.
The same model can predict properties directly for the points of a reconstruction, without having to fuse separate per-view predictions, or produce per-view maps by querying every pixel, which then agree across views as far as their correspondences do.
\method also benefits directly from more accurate geometry and correspondences, such as those of a dense matcher, without retraining.
In~\cref{fig:teaser}, \method predicts the intrinsic properties of the points of a laser-scanned scene~\cite{schops2017multi}, so that a light placed once or a material edited once produces consistent results across the input views.

We evaluate \method in three experiments.
First, with the geometry of the scene given, we query the surface points of rendered scenes~\cite{roberts2021hypersim, zhu2022learning} directly: \method is competitive with or more accurate than baselines whose per-view predictions are averaged at each point, and has no disagreement between views to resolve.
Second, we measure the consistency of per-view maps at exact correspondences on laser-scanned real scenes~\cite{schops2017multi}, where \method is the most consistent.
Third, we measure prediction accuracy on rendered scenes and real photographs~\cite{wu2023measured}: \method is competitive with the state of the art from a single image, and becomes more accurate with every added image.
Ablations show that pooling the features of all views into a single representation per point brings this consistency at no cost in accuracy: decoding every pixel from its own features is equally accurate, but far less consistent.

In summary, our contributions are: (1) a formulation that predicts intrinsic properties once per surface point rather than per pixel, and is consistent across views by construction; (2) \method, a feed-forward model that implements it from one or several unposed images, and serves both per-point and per-view prediction; and (3) an evaluation of intrinsics on surface points, and of their consistency at exact correspondences on real scenes.

\begin{figure*}[t]
    \centering
    \includegraphics{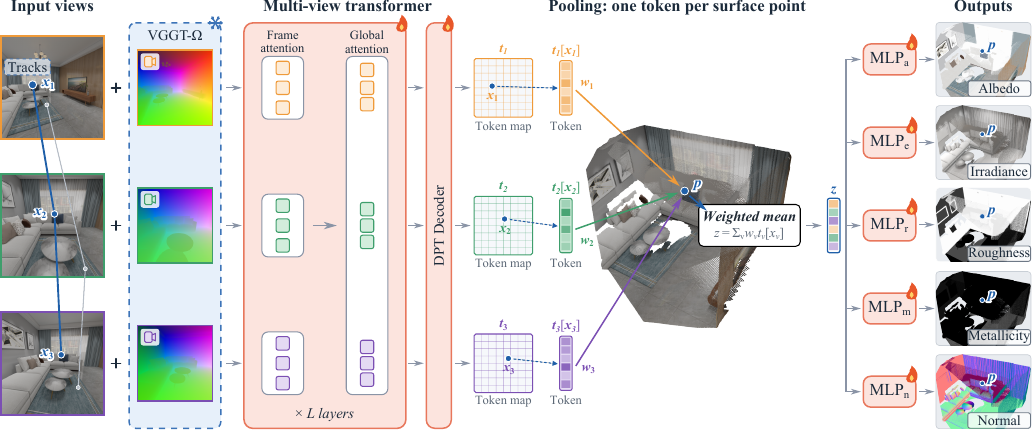}
    \caption{\methodbf \textbf{overview}. A multi-view transformer, conditioned on the geometry estimated by frozen VGGT-$\Omega$~\cite{wang2026vggtomega}, turns each view $v$ into a token map $t_v$. For a queried surface point $p$, the tokens at its locations $x_v$ in the views are pooled into a single token $z$ by a confidence-weighted mean, and one MLP per property decodes the point's intrinsics from $z$, consistent across views by construction.}
    \label{fig:method}
\end{figure*}
\section{Related Work}
\label{sec:related}

\paragraph{Intrinsic decomposition and inverse rendering.}
Intrinsic images~\cite{barrow1978recovering, land1971lightness} separate an image into reflectance and shading. Early methods relied on physical assumptions and hand-designed priors~\cite{horn1974determining, weiss2001deriving, tappen2002recovering, grosse2009ground, barron2014shape}; later ones learned priors from human judgments~\cite{bell2014intrinsic, zhou2015learning, narihira2015direct}, rendered scenes~\cite{li2018cgintrinsics, roberts2021hypersim, zhu2022learning} and time-lapse video~\cite{li2018learning}. Inverse rendering widens the output to materials and lighting for whole indoor scenes~\cite{sengupta2019neural, li2020inverse, wang2021learning, zhu2022irisformer, zhu2022learning}. Generative models have recently raised single-image accuracy~\cite{du2023generative, kocsis2024intrinsic, zeng2024rgb, sartor2025teamwork, careaga2024colorful, ke2025marigold, pavlovic2026marigoldv2, chen2024intrinsicanything, sun2026unid}, since they can resolve the ambiguity with a plausible sample. All of these methods predict single-view, per-pixel intrinsics.

\paragraph{Consistency across views and time.}
The disagreement between views has been addressed in three ways. Intrinsic video and multi-view intrinsic images tie the reflectance along correspondences with a penalty, solved per sequence~\cite{laffont2012coherent, laffont2012rich, kong2014intrinsic, ye2014intrinsic, fischer2026sama, bonneel2014interactive, duchene2015multi, meka2016live}; later work used such sequences as a training signal for single-image networks~\cite{li2018learning, lettry2018unsupervised, yu2019inverserendernet}. Reconstruction-based inverse rendering attaches materials and lighting to a 3D representation, optimized per scene~\cite{zhang2021physg, srinivasan2021nerv, boss2021nerd, zhang2021nerfactor, zhang2022modeling, yao2022neilf, jin2023tensoir, zhang2023neilf++, wu2023factorized, gao2024relightable, liang2024gs, jiang2024gaussianshader, niu2026sgs, kocsis2026intrinsic} or predicted by a network~\cite{zhang2025relitlrm, li2025lirm, karpikova2026invsplat, schmid2026genrecon, zhang2026worldrecon}, and renders it into every view; this is consistent by design, but every view inherits its errors. Most recently, multi-view and video networks exchange information across views through attention~\cite{li2025idarb, wu2026mvinverse, du2026mvid, liang2025diffusionrenderer, vrgbx2025}, project predictions between frames~\cite{dirik2026geoid}, or fine-tune with a consistency objective~\cite{wu2026mvinverse}. They are more consistent than single-image networks, but the maps of each view are still decoded independently, so corresponding pixels can disagree. We instead decode the intrinsics once per surface point.

\paragraph{Feed-forward reconstruction, matching and tracking.}
Feed-forward reconstruction models~\cite{wang2024dust3r, leroy2024grounding, wang2025vggt, lin2025depth, karhade2025any4d, wang2026pi, keetha2026mapanything, wang2026vggtomega, luo20264rc, sucar2026vdpm} predict depth and cameras for a set of images or a video with a transformer that attends across views. We use one as our multi-view transformer, and a frozen copy to estimate the geometry that conditions it. Dense matchers~\cite{sun2021loftr, truong2023pdc, edstedt2023dkm, edstedt2024roma, edstedt2026roma, zhang2025ufm, nordstrom2026romaomega, lindenberger2026flexglue} estimate a correspondence and its certainty for every pixel of an image pair without recovering 3D geometry, and point trackers~\cite{harley2022particle, doersch2022tap, doersch2023tapir, karaev2024cotracker, karaev2025cotracker3, singer2026reliftdense3dtracking, lai2026cowtracker} follow a pixel's surface point through a video. In our framework, a track is the pixel at which one surface point is visible in each view, obtained by projecting a 3D point with this geometry or by matching a pixel across the images, which is more precise and what our per-view maps rely on.

\paragraph{Pooling the observations of a point.}
Pooling per-view features at a 3D point is a staple of generalizable view synthesis and reconstruction: pixelNeRF~\cite{yu2021pixelnerf} averages them, IBRNet~\cite{wang2021ibrnet} and NeuRay~\cite{liu2022neuray} weight them by learned visibility, MVSNeRF~\cite{chen2021mvsnerf} fuses them in a cost volume, transformers attend along epipolar lines~\cite{suhail2022generalizable} or over all view tokens without an explicit point~\cite{sajjadi2022scene, hong2024lrm, zhang2024gslrm, jin2025lvsm}, and GenRecon~\cite{schmid2026genrecon} conditions a generative mesh model on their statistics. Our pooling follows these works: a visibility-weighted mean of the tokens of a surface point, decoded into intrinsic properties. A second line lifts per-view features or labels into a 3D field and consolidates their disagreements there~\cite{zhi2021place, kobayashi2022decomposing, tschernezki2022neural, kerr2023lerf, siddiqui2023panoptic, fischer2024nerf, zhou2024feature, marrie2025ludvig, fischer2026sama}, at the cost of a per-scene optimization. We pool the tokens before decoding, so there are no per-view predictions to consolidate.
\section{Method}
\label{sec:method}

We introduce \method, illustrated in~\cref{fig:method}, a feed-forward transformer that takes a set of images of a scene and predicts the intrinsic properties of any queried surface point.
\Cref{sec:framework} states the problem and how a surface point is located in the images,~\cref{sec:architecture} describes the architecture, and~\cref{sec:training} the training setup.

\subsection{Querying Surface Points}
\label{sec:framework}

\paragraph{Problem.}
Let $I_1,\dots,I_N\in\mathbb{R}^{3\times H\times W}$ be $N\geq 1$ images of a scene and $p\in\mathbb{R}^3$ a point on one of its surfaces.
\method predicts the intrinsic properties of $p$,
\begin{equation}
  (a, e, r, m, n) = f\big(p;\, I_1,\dots,I_N\big),
  \label{eq:problem}
\end{equation}
where $a\in[0,1]^3$ is the albedo, $e\in\mathbb{R}^3_{\geq 0}$ the irradiance, $r\in[0,1]$ the roughness, $m\in[0,1]$ the metallicity and $n\in\mathbb{S}^2$ the normal.
These are properties of 3D surface points instead of images, and thus should be identical across views.
We therefore let the surface point, rather than a pixel of an image, be the unit of prediction.
Since $f$ depends only on the point and the images, and on no reference view, every surface point receives a single prediction, consistent across views by construction.

\paragraph{Geometry.}
A 3D point is only meaningful relative to the geometry of the scene.
We assume a camera and a depth map $D_v$ for every view $v$.
They can be given, for example by a scan or a multi-view reconstruction, or, for unposed images, estimated from the images themselves by a reconstruction model.
Unprojecting $D_v$ yields the point map $X_v$, the 3D point observed at each pixel of view $v$.

\paragraph{Tracks.}
To predict the properties of $p$, we first locate it in the input images.
The \emph{track} of $p$ is the set of locations $x_v$ at which it appears in the views, which we obtain by projecting $p$ with the camera of each view $v$.
Projection returns a location even where $p$ is occluded, so each location $x_v$ carries a confidence $c_v\in[0,1]$ that $p$ is visible in view $v$.
We estimate it from the distance between $p$ and the surface point imaged at $x_v$,
\begin{equation}
    c_v = \frac{1}{1 +\delta_v/\tau},
    \qquad
    \delta_v = \frac{\lVert X_v[x_v] - p \rVert}{d_v},
    \label{eq:confidence}
\end{equation}
where $X_v[x_v]$ is the point map sampled bilinearly at $x_v$, $d_v$ is the depth of $p$ in view $v$, and $\tau$ is a relative tolerance, which we set to $1\%$.
If $p$ lies behind the camera ($d_v \leq 0$) or does not project into the image, $c_v = 0$.
Where $p$ is visible, $\delta_v\approx 0$ and the confidence is close to one.
The farther the imaged surface lies from $p$, the lower the confidence.
\method needs only the track of a query, so its locations can equally come from any other source of correspondences.

\paragraph{Pixel queries.}
A pixel query is a special case of~\cref{eq:problem}: a pixel $q$ of view $i$ maps to a single surface point, $p = X_i[q]$.
Querying every pixel of an image yields its intrinsic maps, while querying the points of a point cloud or the texels of a mesh yields them directly on that representation.
The track of a pixel can be obtained by projection or directly in image space with a dense matcher such as RoMa-$\Omega$~\cite{nordstrom2026romaomega}; $c_v$ of~\cref{eq:confidence}, still computed with the point maps and cameras at $x_v$, is then multiplied by the dense matcher's confidence.

\begin{figure*}[htbp]
    \centering
    \includegraphics{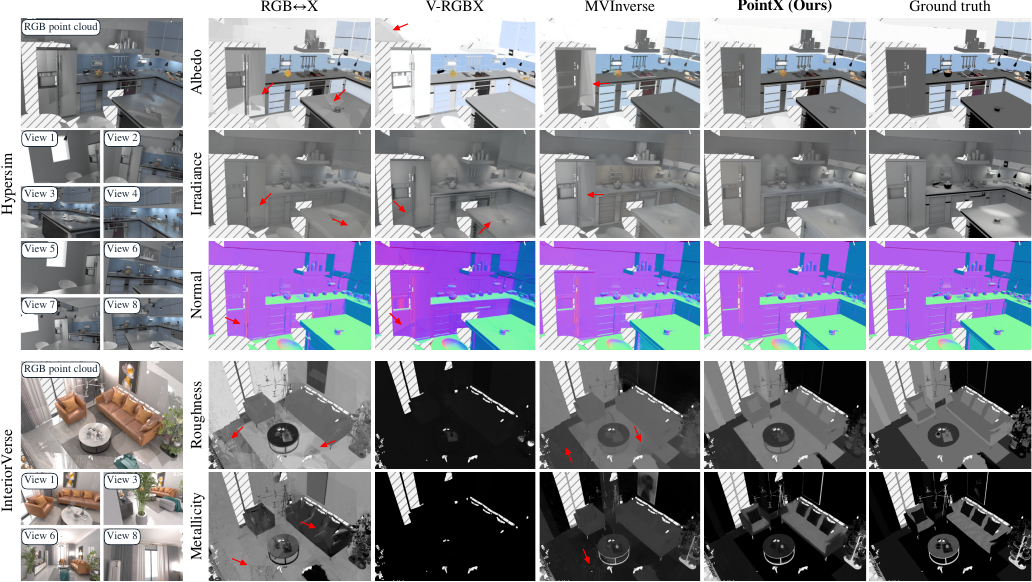}
    \caption{\textbf{Intrinsics of surface points.} Point clouds of a Hypersim~\cite{roberts2021hypersim} (top) and an InteriorVerse~\cite{zhu2022learning} (bottom) scene, built from the ground-truth depth of the input views and rendered from a new viewpoint (regions without ground truth are hatched).
    The baselines average their per-view maps over the views in which a point is visible; where these views disagree, the average shows seams and patches (red arrows).
    \method predicts every point once and needs no averaging, yielding uniform surfaces.}
    \label{fig:qual_surface}
\end{figure*}

\subsection{Architecture}
\label{sec:architecture}

\method consists of three parts (\cref{fig:method}): a multi-view transformer turns every image into a map of tokens, a pooling step reads one token per view along the track of the query and averages them into a single token, and MLP heads decode the intrinsic properties of the point from this token.

\paragraph{Input geometry.}
We run VGGT-$\Omega$~\cite{wang2026vggtomega} (frozen) once on the input images to estimate a camera and a depth map with per-pixel confidence for every view.
This geometry is an input of our multi-view transformer, but we first change the frame in which it is expressed: VGGT-$\Omega$ expresses its camera poses relative to the first view, whereas our model should treat all views symmetrically.
We therefore express all geometry in a common frame, the \emph{mean-frame}: its origin is the mean of the camera centers, its forward axis the normalized mean of the viewing directions, its horizontal axis the mean of the cameras' horizontal axes made orthogonal to the forward axis, and its vertical axis the cross product of the two.
No view is privileged in this frame, and for a single view it coincides with the camera frame.
The predicted normal $n$ is expressed in the mean-frame as well.

\paragraph{Multi-view transformer.}
The transformer computes one feature map per view from all images jointly, so that the features of a view combine evidence from all views.
We use the transformer of VGGT-$\Omega$~\cite{wang2026vggtomega}, initialized from its released weights and fine-tuned, which alternates self-attention within each view (frame attention) and across all views (global attention)~\cite{wang2025vggt}.
Since a normal is decoded from a token pooled over several views, the features of every view must encode the pose of its camera in the mean-frame, which the image alone cannot provide.
We therefore give every pixel of a view ten geometry channels, all expressed in the mean-frame: its viewing ray in 6D Pl\"ucker coordinates~\cite{sitzmann2021light}, the 3D point that VGGT-$\Omega$ estimates at that pixel, and the scalar log-confidence of this estimate.
An additional, zero-initialized patch embedding embeds these channels, and its output is added to the patch embedding of the image, so that the pretrained features are unchanged at initialization.
Further, VGGT-$\Omega$ marks its first view with separate camera and register tokens.
We drop the camera tokens and use the same register tokens for every view, making our transformer, unlike the frozen VGGT-$\Omega$ that estimates its input geometry, equivariant to the order of views.

\paragraph{Token map.}
The transformer outputs feature maps at $1/16$ of the image resolution, too coarse for per-pixel detail, which a DPT decoder~\cite{ranftl2021vision} upsamples to the full image resolution (see~\cref{sec:app:arch}).
The \emph{token map} $t_v$ of view $v$ then holds one token per pixel, built from two sources.
The first is a linear projection of the upsampled feature at that pixel, which carries the multi-view evidence of the transformer.
Inspired by D4RT~\cite{zhang2026d4rt}, we add a linear projection of the $9\times 9$ RGB window of $I_v$ around the pixel.
Both projections are shared by all views.

\paragraph{Pooling.}
To decode the intrinsics of $p$ once, from all the views in which it is visible, rather than once per view, we merge the tokens along its track into a single token.
Every view contributes the token at its location on the track, $t_v[x_v]$, sampled bilinearly, and we average these tokens with their confidence values along the track as weights,
\begin{equation}
  z = \sum_{v} w_v\, t_v[x_v],
  \qquad
  w_v = \frac{c_v}{\sum_{u} c_u}.
  \label{eq:pool}
\end{equation}
A view in which $p$ is occluded, or in which $x_v$ falls on the background beside a thin structure, has a low confidence and contributes little to $z$.
The weights have no learned parameters and depend only on the point and the views.
For a pixel query, the queried view $i$ has $\delta_i = 0$ and thus full confidence.
The weighted mean is invariant to the order of the views and defined for any number of them, provided $\sum_u c_u > 0$; for a single image, $z$ is simply the token at $x_1$.

\paragraph{Heads.}
Each intrinsic is decoded from $z$ by an individual two-layer MLP, followed by a sigmoid for albedo, roughness and metallicity, a softplus for irradiance which is predicted directly as $\log(1+e)$, and a unit-length normalization for the normal.
To obtain the normal map of a view, we rotate $n$ from the mean-frame into that view's camera frame.

\paragraph{Consistency.}
Nothing in~\cref{eq:pool} refers to the view from which a point is queried.
With a track obtained by projection, the locations, the weights and hence $z$ depend only on the point, the geometry and the images.
A surface point thus receives a single prediction from whichever view it is queried: its disagreement across views is zero by construction, without a consistency loss or fusion of per-view predictions.
Per-view maps are consistent up to a small disagreement, because corresponding pixels of two views do not query exactly the same point.
Even with exact geometry, the two pixels unproject to neighboring points, since each view samples the surface on its own pixel grid.
Errors in the estimated depth and cameras move them further apart and tilt the normals.
With a matcher, the two tracks also differ slightly, as each is matched from its own view.

\begin{table*}[tb!]
\centering
\small
\tabcolsep=0.45em
\caption{\textbf{Intrinsics of surface points.} Accuracy of the per-point predictions, and disagreement between a method's per-view predictions before they are averaged; \method predicts each point once, so its disagreement is zero by construction (not ranked). The first row applies the same protocol to the ground-truth maps, each rendered on its own pixel grid. PSNR$_\text{s}$: PSNR after a per-channel scale; RMSE$_\text{r}$: RMSE between two views relative to random pairs. Best, second and third best are highlighted. $^\dagger$Trained on the Hypersim test scenes, not ranked.}
\resizebox{\linewidth}{!}{
\begin{tabular}{@{}l cccc GGG ccccc GGGG@{}}
\toprule
\multirow{4}[6]{*}{\textbf{Method}} & \multicolumn{7}{c}{\textbf{Hypersim}~\cite{roberts2021hypersim}} & \multicolumn{9}{c}{\textbf{InteriorVerse}~\cite{zhu2022learning}} \\
\cmidrule(lr){2-8}\cmidrule(lr){9-17}
 & \multicolumn{4}{c}{\textbf{Accuracy}} & \multicolumn{3}{G}{\textbf{Disagreement}} & \multicolumn{5}{c}{\textbf{Accuracy}} & \multicolumn{4}{G}{\textbf{Disagreement}} \\
\cmidrule(lr){2-5}\cmidrule(lr){6-8}\cmidrule(lr){9-13}\cmidrule(lr){14-17}
 & \multicolumn{2}{c}{\textbf{Albedo}} & \textbf{Irrad.} & \textbf{Normal} & \textbf{Albedo} & \textbf{Irrad.} & \textbf{Normal} & \multicolumn{2}{c}{\textbf{Albedo}} & \textbf{Normal} & \textbf{Rough.} & \textbf{Metal.} & \textbf{Albedo} & \textbf{Normal} & \textbf{Rough.} & \textbf{Metal.} \\
\cmidrule(lr){2-3}\cmidrule(lr){4-4}\cmidrule(lr){5-5}\cmidrule(lr){6-6}\cmidrule(lr){7-7}\cmidrule(lr){8-8}\cmidrule(lr){9-10}\cmidrule(lr){11-11}\cmidrule(lr){12-12}\cmidrule(lr){13-13}\cmidrule(lr){14-14}\cmidrule(lr){15-15}\cmidrule(lr){16-16}\cmidrule(lr){17-17}
 & PSNR~$\uparrow$ & PSNR$_\text{s}$~$\uparrow$ & PSNR$_\text{s}$~$\uparrow$ & Mean$^\circ$~$\downarrow$ & RMSE$_\text{r}$~$\downarrow$ & RMSE$_\text{r}$~$\downarrow$ & Mean$^\circ$~$\downarrow$ & PSNR~$\uparrow$ & PSNR$_\text{s}$~$\uparrow$ & Mean$^\circ$~$\downarrow$ & RMSE~$\downarrow$ & RMSE~$\downarrow$ & RMSE$_\text{r}$~$\downarrow$ & Mean$^\circ$~$\downarrow$ & RMSE$_\text{r}$~$\downarrow$ & RMSE$_\text{r}$~$\downarrow$ \\
\midrule
\textit{Ground-truth maps} & 43.47 & 43.53 & 42.58 & 0.7 & 0.089 & 0.145 & 3.1 & 45.12 & 45.14 & 0.6 & 0.006 & 0.008 & 0.093 & 3.1 & 0.083 & 0.173 \\
\midrule
Intrinsic Image Diffusion~\cite{kocsis2024intrinsic} & 12.37 & 14.21 & -- & -- & 0.483 & -- & -- & 16.11 & 19.12 & -- & \thirdbest{0.201} & \thirdbest{0.130} & 0.439 & -- & \thirdbest{0.562} & \secondbest{0.493} \\
RGB$\leftrightarrow$X$^\dagger$~\cite{zeng2024rgb} & \textcolor{gray}{24.09} & \textcolor{gray}{24.89} & \textcolor{gray}{24.17} & \textcolor{gray}{8.0} & \textcolor{gray}{0.196} & \textcolor{gray}{0.289} & \textcolor{gray}{8.3} & 9.57 & 18.96 & 12.3 & 0.304 & 0.407 & 0.499 & 18.1 & 0.582 & 0.837 \\
DiffusionRenderer~\cite{liang2025diffusionrenderer} & 14.50 & 16.82 & -- & 13.6 & 0.312 & -- & 12.2 & 10.82 & 17.92 & 11.4 & 0.292 & 0.192 & 0.513 & 14.0 & 0.821 & 0.954 \\
V-RGBX~\cite{vrgbx2025} & 11.98 & 15.57 & 18.92 & 23.3 & 0.239 & \thirdbest{0.278} & 11.9 & 11.42 & 14.48 & 22.4 & 0.407 & 0.193 & 0.431 & 16.6 & 0.789 & 0.959 \\
MVInverse~\cite{wu2026mvinverse} & 18.62 & \secondbest{26.01} & 19.22 & \best{8.2} & \best{0.141} & 0.302 & 6.4 & 18.41 & 19.33 & \secondbest{7.8} & \secondbest{0.152} & \secondbest{0.111} & \thirdbest{0.218} & \best{8.8} & \best{0.241} & \best{0.316} \\
RGB$\leftrightarrow$X + GeoID$^\dagger$~\cite{dirik2026geoid} & \textcolor{gray}{23.62} & \textcolor{gray}{24.58} & -- & -- & \textcolor{gray}{0.186} & -- & -- & 10.89 & 18.99 & -- & 0.266 & 0.419 & 0.481 & -- & \secondbest{0.553} & \thirdbest{0.815} \\
UniD~\cite{sun2026unid} & 17.34 & 18.65 & \thirdbest{20.72} & 14.9 & 0.189 & 0.283 & 11.0 & 9.63 & 16.26 & 15.7 & -- & -- & 0.341 & 17.7 & -- & -- \\
Marigold V2~\cite{pavlovic2026marigoldv2} & \secondbest{21.16} & 22.64 & -- & \thirdbest{8.4} & 0.209 & -- & \thirdbest{5.9} & 11.57 & 17.47 & \thirdbest{8.7} & -- & -- & 0.382 & \thirdbest{10.5} & -- & -- \\
MVID~\cite{du2026mvid} & \best{26.22} & \best{27.01} & \best{25.19} & 8.9 & \secondbest{0.147} & \best{0.228} & \best{4.9} & \secondbest{24.64} & \secondbest{25.15} & 24.3 & -- & -- & \best{0.157} & 12.7 & -- & -- \\
MVID-$\Omega$~\cite{du2026mvid} & 17.72 & 19.53 & 19.34 & 11.5 & \thirdbest{0.172} & \secondbest{0.252} & \secondbest{5.0} & \thirdbest{19.32} & \thirdbest{21.03} & 25.4 & -- & -- & \secondbest{0.174} & \secondbest{9.0} & -- & -- \\
\midrule
\textit{\method w/o pooling, maps averaged} & 20.33 & 22.78 & 22.66 & 8.6 & 0.136 & 0.212 & 4.1 & 25.41 & 25.89 & 5.7 & 0.083 & 0.071 & 0.158 & 4.8 & 0.144 & 0.244 \\
\textbf{\method (Ours)} & \thirdbest{20.32} & \thirdbest{22.76} & \secondbest{22.69} & \secondbest{8.3} & \multicolumn{3}{c}{\cellcolor{constructgreen}\textbf{0} by construction} & \best{25.50} & \best{25.97} & \best{5.2} & \best{0.083} & \best{0.072} & \multicolumn{4}{c}{\cellcolor{constructgreen}\textbf{0} by construction} \\
\bottomrule
\end{tabular}
}
\label{tab:surface}
\end{table*}

\subsection{Training}
\label{sec:training}

Since the ground truth of our datasets comes as per-view maps, we train with pixel queries.
At each step, we sample query pixels in one view of every scene and supervise their intrinsics with the ground truth.
The tracks of these queries are drawn at random from three sources: projection with the ground-truth or VGGT-$\Omega$'s geometry, and the dense matcher RoMa-$\Omega$~\cite{nordstrom2026romaomega}.
We apply an $L_1$ loss on albedo, roughness, metallicity and log-irradiance, the cosine distance on the normal, and a multi-scale gradient loss~\cite{li2018megadepth, wang2025vggt} on queries sampled as $16\times16$ windows.
As rendered frames are exposed individually, we bring the irradiance labels of a set of views to one exposure, so that all views learn the same irradiance.
We neither compare the predictions of different views nor use visibility labels: the pooling of~\cref{eq:pool} has no learned parameters, so consistency holds by construction, not by learning.
We train on rendered scenes of Hypersim~\cite{roberts2021hypersim}, InteriorVerse~\cite{zhu2022learning} and three in-house datasets (1,400 in- and outdoor scenes), and on $1$M real photographs pseudo-labeled by a single-image model (see~\cref{sec:app:data}).

\section{Experiments}
\label{sec:experiments}

We compare \method with recent state-of-the-art methods in two settings.
In~\cref{sec:surface}, the geometry of the scene is given and we query its surface points directly.
In~\cref{sec:consistency,sec:accuracy}, only the images are given: we query every pixel of every image and evaluate the consistency and the accuracy of the resulting per-view maps.
Finally, we present a set of key ablations in~\cref{sec:ablations} and discuss applications in~\cref{sec:applications}.

\paragraph{Baselines.}
We compare with single-view methods~\cite{kocsis2024intrinsic, zeng2024rgb, pavlovic2026marigoldv2}, video models~\cite{liang2025diffusionrenderer, vrgbx2025, sun2026unid}, the multi-view models MVInverse~\cite{wu2026mvinverse}, MVID and MVID-$\Omega$~\cite{du2026mvid}, and GeoID~\cite{dirik2026geoid}, a test-time multi-view consensus on RGB$\leftrightarrow$X~\cite{zeng2024rgb}.
All use their released weights, and the video and multi-view models process several images jointly.
Methods that need posed images~\cite{schmid2026genrecon} or per-scene optimization~\cite{kocsis2026intrinsic} are not part of the main comparison.
We evaluate GenRecon~\cite{schmid2026genrecon} and Intrinsic Image Fusion~\cite{kocsis2026intrinsic}, on the task of~\cref{sec:surface} in~\cref{sec:app:baselines_app}.
We omit methods that target single objects~\cite{li2025idarb} or have no released model~\cite{karpikova2026invsplat}.

\paragraph{Datasets and metrics.}
Every dataset is evaluated on sets of up to eight views of a scene.
Hypersim~\cite{roberts2021hypersim} and InteriorVerse~\cite{zhu2022learning} are rendered and come with ground-truth intrinsics and depth, MAW~\cite{wu2023measured} provides measured albedo for real indoor photographs, and ETH3D~\cite{schops2017multi} provides real photographs with laser-scanned geometry.
For accuracy, we report the PSNR of the albedo, as predicted and after fitting a per-channel scale to the ground truth (PSNR$_\text{s}$); the PSNR$_\text{s}$ of the irradiance, which is defined only up to the exposure of an image; the mean angular error of the normals; and the RMSE of roughness and metallicity.
The disagreement between two views is measured at corresponding points: we report the mean angle between their normals and, for the other properties, the RMSE between corresponding points divided by the RMSE between randomly paired ones (RMSE$_\text{r}$, see appendix), so that a constant map is not rewarded.
An RMSE$_\text{r}$ of $0$ means that the views agree, and an RMSE$_\text{r}$ of $1$ that they agree no better than at random.

\begin{figure*}[htbp]
    \centering
    \includegraphics{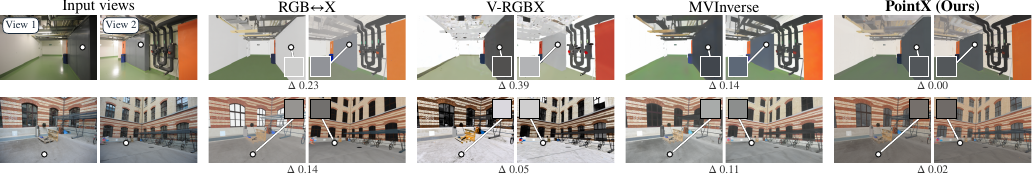}
    \caption{\textbf{Consistency of per-view maps on ETH3D~\cite{schops2017multi}.} Albedo predictions for two views of a scene (\method: RoMa-$\Omega$ tracks). The marked surface point is the same in both views; the insets show the value each method predicts for it, and $\Delta$ their mean absolute difference.}
    \label{fig:qual_consistency}
\end{figure*}

\subsection{Intrinsics of Surface Points}
\label{sec:surface}

We first evaluate the task of~\cref{eq:problem}: predicting the intrinsics of a surface point from all the images in which it is visible.

\paragraph{Protocol.}
We use Hypersim and InteriorVerse, whose ground-truth depth gives the surface of every scene.
InteriorVerse provides no camera poses, so we register its views from the depth maps (see~\cref{sec:app:eval}).
For every scene, we unproject the depth maps and sample surface points that are visible from at least two views.
We define the true value of a point to be the median of the ground-truth maps at its projections.
\method is queried once per point, with its track and the confidence of~\cref{eq:confidence} obtained by projection with the ground-truth geometry.
The baselines predict one map per view, so we read these maps at the projections of the point and take their unweighted mean over the views in which it is visible, using the same geometry; a weighted mean or a median gives similar results (see~\cref{sec:app:surface_app}).

\paragraph{Results.}
\Cref{tab:surface} shows that \method is the most accurate on every property of InteriorVerse and competitive with the most accurate baselines on Hypersim.
Averaging the maps that \method predicts for each view without pooling gives nearly the same accuracy, but at the cost of consistency.
Every baseline assigns different values to the same surface point in different views, so its predictions must be fused before they can be attached to the scene.
As observed in~\cref{fig:qual_surface}, where there is disagreement, seams appear on the baselines' intrinsic point clouds.
\method decodes the intrinsics of every point once, which yields uniform surfaces.

\subsection{Consistency of Per-View Maps}
\label{sec:consistency}

The geometry of a scene is often unknown, and many applications expect one map per image.
We hence measure next whether a method's per-view maps agree at ground-truth correspondences.

\paragraph{Protocol.}
We apply the reprojection protocol of MVInverse~\cite{wu2026mvinverse} to the real photographs of ETH3D, whose laser scans give exact correspondences.
For every pair of views, the pixels of one view are projected into the other with the ground-truth depth and kept where the depth of the other view confirms that they are visible.
The maps of the two views are compared at these corresponding pixels, with the normals rotated into the world frame by the ground-truth cameras.
Additional results are in the appendix.

\paragraph{Results.}
\Cref{tab:consistency} shows that \method is the most consistent on all five properties with the tracks of a dense matcher, RoMa-$\Omega$.
In~\cref{fig:qual_consistency}, the marked surface point looks different in the two views, and the baselines assign it a different albedo in each view, whereas \method's predictions look the most similar and have the lowest difference.
The video and multi-view baselines process the views jointly but still decode each view separately, which does not ensure consistency.
In \method, corresponding pixels are decoded from tokens pooled along nearly identical tracks, so their predictions can differ only as much as their tracks do (\cref{sec:architecture}).
Its consistency therefore improves with the precision of the tracks, without retraining.
With tracks projected through the coarser geometry of VGGT-$\Omega$, it already leads on irradiance, normals and roughness.
The matcher extends the lead to all four properties.
When ground-truth geometry is available, its exact tracks make \method highly consistent, with by far the lowest disagreement on every property.

\begin{table}[htbp]
\centering
\small
\tabcolsep=0.5em
\caption{\textbf{Consistency of per-view maps on ETH3D~\cite{schops2017multi}.} Disagreement between the maps of two views at ground-truth correspondences. Gray: ground-truth geometry at test time, not ranked. Best, second and third best are highlighted.}
\resizebox{\columnwidth}{!}{
\begin{tabular}{@{}l c ccccc@{}}
\toprule
\multirow{2}[2]{*}{\textbf{Method}} & \multirow{2}[2]{*}{\textbf{Track Source}} & \textbf{Albedo} & \textbf{Irrad.} & \textbf{Normal} & \textbf{Rough.} & \textbf{Metal.} \\
\cmidrule(lr){3-3}\cmidrule(lr){4-4}\cmidrule(lr){5-5}\cmidrule(lr){6-6}\cmidrule(lr){7-7}
 & & RMSE$_\text{r}$~$\downarrow$ & RMSE$_\text{r}$~$\downarrow$ & Mean$^\circ$~$\downarrow$ & RMSE$_\text{r}$~$\downarrow$ & RMSE$_\text{r}$~$\downarrow$ \\
\midrule
\multicolumn{7}{@{}l}{\textit{Single-view methods}} \\
Intrinsic Image Diffusion~\cite{kocsis2024intrinsic} & -- & 0.481 & -- & -- & 0.759 & 0.581 \\
RGB$\leftrightarrow$X~\cite{zeng2024rgb} & -- & 0.378 & 0.448 & 13.5 & 0.562 & 0.799 \\
Marigold V2~\cite{pavlovic2026marigoldv2} & -- & 0.210 & -- & 7.4 & -- & -- \\
\midrule
\multicolumn{7}{@{}l}{\textit{Video \& Multi-view methods}} \\
DiffusionRenderer~\cite{liang2025diffusionrenderer} & -- & 0.291 & -- & 11.6 & 0.568 & 0.733 \\
V-RGBX~\cite{vrgbx2025} & -- & 0.216 & \thirdbest{0.302} & 10.4 & \thirdbest{0.334} & 0.870 \\
MVInverse~\cite{wu2026mvinverse} & -- & \secondbest{0.161} & 0.318 & 9.6 & 0.382 & \secondbest{0.270} \\
RGB$\leftrightarrow$X + GeoID~\cite{dirik2026geoid} & -- & 0.374 & -- & -- & 0.556 & 0.783 \\
UniD~\cite{sun2026unid} & -- & 0.200 & 0.327 & 11.1 & -- & -- \\
MVID~\cite{du2026mvid} & -- & 0.202 & 0.335 & 11.0 & -- & -- \\
MVID-$\Omega$~\cite{du2026mvid} & -- & 0.181 & 0.303 & \thirdbest{6.8} & -- & -- \\
\multirow{3}{*}{\textbf{\method (Ours)}} & VGGT-$\Omega$~\cite{wang2026vggtomega} & \thirdbest{0.177} & \secondbest{0.214} & \secondbest{4.6} & \secondbest{0.217} & \thirdbest{0.353} \\
 & RoMa-$\Omega$~\cite{nordstrom2026romaomega} & \best{0.095} & \best{0.146} & \best{2.9} & \best{0.136} & \best{0.222} \\
 & Ground truth & \textcolor{gray}{0.058} & \textcolor{gray}{0.079} & \textcolor{gray}{1.0} & \textcolor{gray}{0.078} & \textcolor{gray}{0.125} \\
\bottomrule
\end{tabular}
}
\label{tab:consistency}
\end{table}

\subsection{Accuracy of Per-View Maps}
\label{sec:accuracy}
\begin{table*}[tb!]
\centering
\small
\tabcolsep=0.4em
\caption{\textbf{Accuracy of per-view maps.} Top: from one image. Bottom: video and multi-view models on several images (\method: RoMa-$\Omega$ tracks). MAW intensity $\times 10^{2}$. Best, second and third best of each block are highlighted. $^\dagger$Trained on the Hypersim test scenes, not ranked.}
\resizebox{\linewidth}{!}{
\begin{tabular}{@{}l ccc c c c ccc c c ccc@{}}
\toprule
\multirow{3}[3]{*}{\textbf{Method}} & \multicolumn{6}{c}{\textbf{InteriorVerse}~\cite{zhu2022learning}} & \multicolumn{5}{c}{\textbf{Hypersim}~\cite{roberts2021hypersim}} & \multicolumn{3}{c}{\textbf{MAW}~\cite{wu2023measured}} \\
\cmidrule(lr){2-7}\cmidrule(lr){8-12}\cmidrule(lr){13-15}
 & \multicolumn{3}{c}{\textbf{Albedo}} & \textbf{Normal} & \textbf{Rough.} & \textbf{Metal.} & \multicolumn{3}{c}{\textbf{Albedo}} & \textbf{Irrad.} & \textbf{Normal} & \multicolumn{3}{c}{\textbf{Albedo}} \\
\cmidrule(lr){2-4}\cmidrule(lr){5-5}\cmidrule(lr){6-6}\cmidrule(lr){7-7}\cmidrule(lr){8-10}\cmidrule(lr){11-11}\cmidrule(lr){12-12}\cmidrule(lr){13-15}
 & PSNR~$\uparrow$ & PSNR$_\text{s}$~$\uparrow$ & LPIPS~$\downarrow$ & Mean$^\circ$~$\downarrow$ & RMSE~$\downarrow$ & RMSE~$\downarrow$ & PSNR~$\uparrow$ & PSNR$_\text{s}$~$\uparrow$ & LPIPS~$\downarrow$ & PSNR$_\text{s}$~$\uparrow$ & Mean$^\circ$~$\downarrow$ & Int.~$\downarrow$ & Chrom.~$\downarrow$ & WHDR~$\downarrow$ \\
\midrule
\multicolumn{15}{@{}l}{\textit{One image}} \\
Intrinsic Image Diffusion~\cite{kocsis2024intrinsic} & 15.84 & 19.02 & \thirdbest{0.163} & -- & \thirdbest{0.211} & \thirdbest{0.141} & 11.67 & 14.71 & 0.455 & -- & -- & 1.36 & 4.16 & 0.212 \\
RGB$\leftrightarrow$X$^\dagger$~\cite{zeng2024rgb} & 9.78 & 18.22 & 0.320 & 19.3 & 0.317 & 0.412 & \textcolor{gray}{22.92} & \textcolor{gray}{24.07} & \textcolor{gray}{0.143} & \textcolor{gray}{23.90} & \textcolor{gray}{11.5} & 0.70 & 4.08 & 0.219 \\
DiffusionRenderer~\cite{liang2025diffusionrenderer} & 13.36 & 18.74 & 0.281 & 10.8 & 0.281 & 0.199 & 12.64 & 16.84 & 0.385 & -- & 15.1 & 0.42 & \thirdbest{3.44} & \best{0.170} \\
V-RGBX~\cite{vrgbx2025} & 9.78 & 15.57 & 0.434 & 20.7 & 0.566 & 0.197 & 12.57 & 16.06 & 0.393 & 17.88 & 22.9 & 0.52 & 4.67 & 0.263 \\
MVInverse~\cite{wu2026mvinverse} & 18.02 & 19.00 & 0.177 & \secondbest{8.7} & \secondbest{0.155} & \secondbest{0.117} & 17.05 & \secondbest{24.28} & 0.240 & 19.17 & \thirdbest{10.6} & \thirdbest{0.34} & 3.93 & 0.223 \\
UniD~\cite{sun2026unid} & 8.74 & 14.28 & 0.351 & 12.4 & -- & -- & 17.11 & 18.75 & 0.352 & \thirdbest{20.08} & 14.3 & 0.36 & 4.77 & 0.229 \\
Marigold V2~\cite{pavlovic2026marigoldv2} & 11.84 & 17.27 & 0.222 & \thirdbest{9.7} & -- & -- & \secondbest{19.90} & \thirdbest{21.92} & \secondbest{0.220} & -- & \best{10.3} & \best{0.24} & \best{3.07} & 0.233 \\
MVID~\cite{du2026mvid} & \secondbest{23.54} & \secondbest{24.31} & \secondbest{0.121} & 11.3 & -- & -- & \best{25.06} & \best{25.83} & \thirdbest{0.235} & \best{23.91} & \secondbest{10.5} & 0.42 & 3.71 & 0.290 \\
MVID-$\Omega$~\cite{du2026mvid} & \thirdbest{18.93} & \thirdbest{20.57} & 0.170 & 11.4 & -- & -- & 17.21 & 19.45 & 0.372 & 19.39 & 13.3 & 0.37 & 3.67 & \secondbest{0.203} \\
\textbf{\method (Ours)} & \best{24.09} & \best{24.87} & \best{0.101} & \best{6.3} & \best{0.100} & \best{0.085} & \thirdbest{18.87} & 21.25 & \best{0.192} & \secondbest{21.25} & 12.3 & \secondbest{0.29} & \secondbest{3.10} & \thirdbest{0.208} \\
\midrule
\multicolumn{15}{@{}l}{\textit{Several images of a scene (up to eight)}} \\
DiffusionRenderer~\cite{liang2025diffusionrenderer} & 10.92 & 17.51 & 0.365 & \thirdbest{13.1} & 0.293 & \thirdbest{0.196} & 13.85 & 16.76 & 0.379 & -- & 16.5 & 0.44 & \thirdbest{3.56} & \best{0.186} \\
V-RGBX~\cite{vrgbx2025} & 11.50 & 14.41 & 0.389 & 24.1 & 0.395 & 0.197 & 11.80 & 16.03 & 0.397 & 18.23 & 24.8 & 0.63 & 4.94 & 0.300 \\
MVInverse~\cite{wu2026mvinverse} & 18.50 & 19.39 & 0.173 & \secondbest{8.6} & \secondbest{0.152} & \secondbest{0.115} & 17.23 & \secondbest{24.61} & \thirdbest{0.237} & 19.19 & \best{10.4} & \secondbest{0.31} & 3.82 & 0.233 \\
RGB$\leftrightarrow$X + GeoID$^\dagger$~\cite{dirik2026geoid} & 11.16 & 18.27 & 0.279 & -- & \thirdbest{0.277} & 0.423 & \textcolor{gray}{21.99} & \textcolor{gray}{23.59} & \textcolor{gray}{0.151} & -- & -- & 0.78 & 3.90 & 0.216 \\
UniD~\cite{sun2026unid} & 9.99 & 16.38 & 0.318 & 17.9 & -- & -- & 16.51 & 18.44 & 0.426 & \thirdbest{19.76} & 17.3 & 0.41 & 4.76 & 0.237 \\
MVID~\cite{du2026mvid} & \secondbest{24.23} & \secondbest{24.89} & \secondbest{0.118} & 25.8 & -- & -- & \best{25.60} & \best{26.22} & \secondbest{0.230} & \best{24.14} & \secondbest{10.7} & 0.41 & 3.58 & 0.295 \\
MVID-$\Omega$~\cite{du2026mvid} & \thirdbest{19.15} & \thirdbest{20.81} & \thirdbest{0.168} & 26.4 & -- & -- & \thirdbest{17.40} & 19.64 & 0.370 & 19.46 & 13.8 & \thirdbest{0.35} & \secondbest{3.54} & \secondbest{0.208} \\
\textbf{\method (Ours)} & \best{25.13} & \best{25.71} & \best{0.103} & \best{5.7} & \best{0.084} & \best{0.077} & \secondbest{19.65} & \thirdbest{22.02} & \best{0.191} & \secondbest{21.42} & \secondbest{10.7} & \best{0.30} & \best{2.87} & \thirdbest{0.214} \\
\bottomrule
\end{tabular}
}
\label{tab:perview}
\end{table*}

We now evaluate the accuracy of per-view maps, predicted from one image and from several images of a scene.

\paragraph{Protocol.}
We evaluate on InteriorVerse, Hypersim and the real photographs of MAW, and additionally report the LPIPS of the albedo.
On MAW, we follow the protocol of~\citet{wu2023measured} and report the intensity, chromaticity and WHDR errors of the albedo.

\begin{figure}[htbp]
\centering
\includegraphics[width=\linewidth]{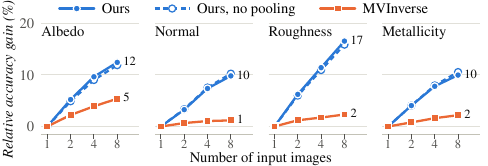}
\caption{\textbf{Accuracy with more input images.} Reduction of the error on InteriorVerse~\cite{zhu2022learning} relative to a single image (RMSE; mean angle for normals). \method uses RoMa-$\Omega$ tracks.}
\label{fig:views}
\end{figure}

\paragraph{Results.}
As shown in~\cref{tab:perview} (top), from a single image, \method is the most accurate on every property of InteriorVerse, and ranks among the top three on most metrics of Hypersim and of MAW, real photographs on which no method was trained.
\Cref{tab:perview} (bottom) further shows that, when given several images of a scene, \method becomes more accurate on all three datasets and leads on most metrics, whereas MVInverse improves only slightly and the video models do not improve consistently.
\Cref{fig:views} shows how the accuracy of a view changes as more views of its scene are added.
Every property predicted by \method becomes more accurate, while MVInverse gains considerably less from the same images.
Most of this gain remains when every pixel is decoded from its own token, without pooling (dashed line): it comes from our multi-view transformer, which gathers more evidence from the additional views.
The same trend holds on the other datasets (see appendix).

\subsection{Ablation Studies}
\label{sec:ablations}

\begin{table}[htbp]
\centering
\small
\tabcolsep=0.4em
\caption{\textbf{Ablation studies} on per-view maps. (a) Without one component, from one and eight images ($^\ast$not retrained). (b) With and without the confidence, from eight images. Best in bold.}
\resizebox{\columnwidth}{!}{
\begin{tabular}{@{}l cc cc cccc@{}}
\toprule
\multicolumn{9}{@{}l}{\textit{(a) Without one component (RoMa-$\Omega$ tracks)}} \\
 & \multicolumn{4}{c}{\textbf{Accuracy} (InteriorVerse~\cite{zhu2022learning})} & \multicolumn{4}{c}{\textbf{Disagreement} (ETH3D~\cite{schops2017multi})} \\
\cmidrule(lr){2-5}\cmidrule(lr){6-9}
 & \multicolumn{2}{c}{\textbf{Albedo} PSNR~$\uparrow$} & \multicolumn{2}{c}{\textbf{Normal} Mean$^\circ$~$\downarrow$} & \textbf{Albedo} & \textbf{Normal} & \textbf{Rough.} & \textbf{Metal.} \\
\cmidrule(lr){2-3}\cmidrule(lr){4-5}\cmidrule(lr){6-6}\cmidrule(lr){7-7}\cmidrule(lr){8-8}\cmidrule(lr){9-9}
 & 1 image & 8 images & 1 image & 8 images & RMSE$_\text{r}$~$\downarrow$ & Mean$^\circ$~$\downarrow$ & RMSE$_\text{r}$~$\downarrow$ & RMSE$_\text{r}$~$\downarrow$ \\
\midrule
\textbf{\method (Ours)} & \textbf{24.09} & \textbf{25.13} & 6.29 & 5.68 & \textbf{0.095} & \textbf{2.91} & \textbf{0.136} & \textbf{0.222} \\
w/o pooling$^\ast$ & \textbf{24.09} & 25.12 & 6.29 & \textbf{5.65} & 0.141 & 5.24 & 0.256 & 0.355 \\
w/o input geometry & 22.92 & 23.83 & 7.53 & 7.10 & 0.098 & 3.09 & 0.149 & 0.278 \\
w/o global attention & 23.90 & 24.31 & \textbf{6.22} & 6.13 & 0.118 & 3.92 & 0.248 & 0.316 \\
\midrule
\multicolumn{9}{@{}l}{\textit{(b) Confidence of \cref{eq:confidence}, by source of tracks}} \\
 & \multicolumn{4}{c}{\textbf{Accuracy} (Hypersim~\cite{roberts2021hypersim}, 8 images)} & \multicolumn{4}{c}{\textbf{Disagreement} (ETH3D~\cite{schops2017multi})} \\
\cmidrule(lr){2-5}\cmidrule(lr){6-9}
 & \multicolumn{2}{c}{\textbf{Albedo} PSNR~$\uparrow$} & \multicolumn{2}{c}{\textbf{Normal} Mean$^\circ$~$\downarrow$} & \multicolumn{2}{c}{\textbf{Albedo} RMSE$_\text{r}$~$\downarrow$} & \multicolumn{2}{c}{\textbf{Normal} Mean$^\circ$~$\downarrow$} \\
\cmidrule(lr){2-3}\cmidrule(lr){4-5}\cmidrule(lr){6-7}\cmidrule(lr){8-9}
 & with & without & with & without & with & without & with & without \\
\midrule
Ground truth & \textbf{20.52} & 18.01 & \textbf{9.23} & 12.26 & 0.058 & 0.058 & 0.99 & 0.99 \\
VGGT-$\Omega$~\cite{wang2026vggtomega} & \textbf{19.49} & 16.99 & \textbf{10.60} & 14.87 & \textbf{0.177} & 0.245 & \textbf{4.63} & 5.31 \\
RoMa-$\Omega$~\cite{nordstrom2026romaomega} & \textbf{20.37} & 19.81 & \textbf{9.56} & 10.27 & \textbf{0.095} & 0.096 & 2.91 & \textbf{1.87} \\
\bottomrule
\end{tabular}
}
\label{tab:ablation}
\end{table}

\paragraph{Pooling.}
Removing the pooling step and decoding every pixel from its own token is as accurate, but makes the views disagree notably more, as shown in~\cref{tab:ablation}a. Pooling brings consistency at no cost in accuracy.

\paragraph{Input geometry.}
Without the geometry channels, we observe in~\cref{tab:ablation}a that every property becomes less accurate and less consistent.
The normals degrade the most: they are expressed in the mean-frame, so the model must infer the orientation of each camera from the images alone, and any error in it tilts all the normals of that view (see~\cref{sec:app:ablations_app}).

\paragraph{Global attention.}
With global attention restricted to each view, the views interact only through pooling.
In~\cref{tab:ablation}a, this variant remains as accurate from one image, but improves far less with more images, and its maps disagree more.
Encoding the views jointly thus turns additional views into accuracy gains (\cref{fig:views}) and aligns the tokens of a surface point across views, which pooling alone does not.

\paragraph{Confidence.}
The confidence of~\cref{eq:confidence} down-weights the views in which a point is occluded.
As observed in~\cref{tab:ablation}b, this improves accuracy with every source of tracks.
The gain is largest with projected tracks, where nothing else rejects the occluded views, and smaller with RoMa-$\Omega$, whose own confidence already rejects most of them.

\paragraph{Runtime.}
On an A100 GPU, \method predicts all maps of eight $512\times336$ images in $1.6$\,s with ground-truth tracks, $2.0$\,s with VGGT-$\Omega$ tracks and $5.8$\,s with RoMa-$\Omega$ tracks.

\subsection{Applications}
\label{sec:applications}

\begin{figure}[htbp]
    \centering
    \includegraphics[width=\linewidth]{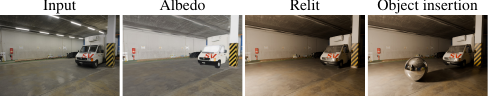}
    \caption{\textbf{Texturing, relighting and object insertion.} The mesh of an ETH3D~\cite{schops2017multi} scene, reconstructed from its laser scans and textured by one \method query per texel, rendered from the camera of the input photograph: its albedo texture, the mesh path traced under a new light with the predicted albedo, roughness, metallicity and normals, and a chrome ball inserted into the relit scene.}
    \label{fig:app_mesh}
\end{figure}

Since \method predicts one set of intrinsics per surface point, its predictions attach directly to a 3D representation of the scene, with no per-view maps to fuse.
In~\cref{fig:teaser}, we query every point of the laser scan of an ETH3D scene once to obtain an intrinsic point cloud.
A new light or an albedo edit applied in 3D then changes every photograph consistently, since each surface point is modified once (see~\cref{sec:app:applications}).
In~\cref{fig:app_mesh}, we instead query one point per texel of a mesh, which bakes the predictions into its textures.
The result is a standard asset for physically based rendering: a path tracer renders it under a new light with the predicted materials, and an inserted chrome ball reflects the textured scene.

\section{Conclusion}
\label{sec:conclusion}

We introduce \method, a simple feed-forward model that predicts the intrinsic properties of any queried surface point from one or more images of a scene.
Our core insight is that intrinsics belong to the surface points of a scene, not to the pixels of an image.
\method therefore pools the evidence of all views in which a point is visible into a single token and decodes its intrinsics once, making each point's prediction consistent across views by construction.
Its per-view maps achieve state-of-the-art consistency on real scenes, and their accuracy, competitive from a single image, improves with every added view to lead on most metrics.
\method also predicts the intrinsics of point clouds and meshes, for view-consistent relighting, editing and object insertion.

\section*{LLM Usage Statement}
We used LLMs to revise parts of the text, and to check grammar and improve the flow of writing.
The authors edited the text, verified its content, and take full responsibility for it. The list of papers to discuss was compiled by the authors together with LLMs.
The authors checked and manually entered all references.
The research direction and ideas came from the authors.
Under the authors' direction, LLMs helped implement the model, training and evaluation code, and set up the baselines and benchmarks from their public repositories.
LLMs were used to check that the method as described in the paper matches the codebase. The authors verified all results and claims.

{
    \small
    \bibliographystyle{ieeenat_fullname}
    \bibliography{main}
}

\clearpage
\appendix
\onecolumn
{
\centering
\Large
\textbf{\thetitle}\\
\vspace{0.5em}Appendix \\
\vspace{1.0em}
}

\section{Model and Training Details}
\label{sec:app:details}

We complement~\cref{sec:method} with further details on the architecture, the training procedure and the training data, in the notation of the main paper.

\subsection{Architecture}
\label{sec:app:arch}

\paragraph{Transformer and decoder.}
The transformer is initialized from the released 1B-parameter checkpoint of VGGT-$\Omega$~\cite{wang2026vggtomega}: a DINOv3 ViT-L/16 patch encoder~\cite{simeoni2025dinov3} followed by 24 frame-attention and 24 global-attention blocks of width 1024 in alternation.
The DPT decoder~\cite{ranftl2021vision} follows the dense head of VGGT-$\Omega$ without its output layer and is trained from scratch.
It fuses the tokens of four transformer blocks into a feature map at $1/4$ of the image resolution, and two further upsampling stages bring it to the full resolution, each with a skip connection from the first two convolution blocks of an ImageNet-pretrained VGG19~\cite{simonyan2014very}.
The decoder feature has 128 channels and the tokens have 256 dimensions.

\paragraph{Tracks.}
For the tracks of RoMa-$\Omega$~\cite{nordstrom2026romaomega}, we use its \texttt{fast} setting and take its predicted overlap probability as the confidence of the matcher.

\paragraph{Heads and model size.}
Each head is a two-layer MLP with a hidden width of 1024 and a GELU between its two layers; its output passes through the activation given in~\cref{sec:architecture}.
In total, \method has 999M trainable parameters: 911M in the transformer, 87M in the decoder and 1.3M in the heads.
The frozen VGGT-$\Omega$ that estimates the input geometry and RoMa-$\Omega$ are not counted.

\subsection{Training}
\label{sec:app:training}

\paragraph{Training sets.}
Each GPU holds 24 images per step, as $\lfloor 24/N\rfloor$ sets of $N$ views, with $N$ drawn from 2 to 24; with probability 0.1, a step instead holds 24 unrelated photographs, each a set of a single view.
A multi-view set consists of frames of one camera trajectory: either a run of frames at a random stride, or a random frame together with frames that are close to it in camera pose.

\paragraph{Augmentation.}
Images are resized and cropped to a width of 512\,px and to a random aspect ratio between 0.33 and 1 that is shared by all images of a step, and each set is mirrored horizontally with probability 0.5.
With probability 0.5, the input images of a set are degraded by a random combination of blur, downsampling, noise and JPEG compression, while their labels are left unchanged.

\paragraph{Queries.}
Each step draws $16{,}384$ query pixels per GPU, all in one randomly chosen view of each set.
Half of them form the $16\times16$ windows of the gradient loss.
We draw $10\%$ of the queries on metallic surfaces, $27\%$ near depth discontinuities and the rest uniformly.
The source of the tracks is drawn uniformly among the three sources of~\cref{sec:training} at every step; where a dataset provides no ground-truth cameras or depth, the geometry of VGGT-$\Omega$ takes the place of the ground truth.

\paragraph{Losses.}
The losses of the five properties are summed with equal weights.
The gradient loss is an $L_1$ loss on the finite differences of prediction and label within each window, at four scales that each subsample the window by two.
Metallic surfaces are rare, so within the metallicity loss, the queries on them are up-weighted by their inverse frequency, capped at 2.
In addition to the queries, the pixels of every view on a grid of stride 4 are decoded from their own token alone and supervised with the same losses, with weight 0.5.
Each loss is applied only where its label is valid: the dataset must provide the property (\cref{tab:app:data}) and define it at the pixel, which excludes, \eg, the normal of the sky and the irradiance of saturated pixels.

\paragraph{Irradiance exposure.}
Where frames are exposed individually, the irradiance label of a surface point differs between two views by the ratio of their exposures.
We estimate this ratio as the median ratio of the labels over the surface points that both views observe, which we find with the ground-truth geometry, solve for the exposure of every view of a set by least squares, and divide its labels by it.
The common exposure of a set remains arbitrary: we fit one scale between prediction and label per set in closed form before the $L_1$ loss, and penalize its logarithm with weight 0.1.
Where the views cannot be related, as for single photographs, the scale is fitted per image.

\paragraph{Optimization.}
We use AdamW~\cite{loshchilov2017decoupled} ($\beta=(0.9,0.95)$, weight decay 0.05), bfloat16 mixed precision, gradient clipping at an $L^2$ norm of 10, and an exponential moving average of the parameters with decay 0.999, which is the model we evaluate.
\method is trained with the losses above for $70$k steps, with $5$k warm-up steps to a learning rate of $1\times10^{-4}$ and a cosine decay to $10^{-5}$.
Training takes approximately $5$ days to complete on 8 A100 80GB GPUs.

\subsection{Training Data}
\label{sec:app:data}

\Cref{tab:app:data} lists the training data.
For Hypersim~\cite{roberts2021hypersim} and InteriorVerse~\cite{zhu2022learning}, we follow the official scene splits and train on the training and validation scenes; for InteriorVerse we use both of its lens versions.
The three in-house datasets contain rendered indoor and outdoor scenes.
The photographs are real images with pseudo-labels from a single-image model and form the single-image steps.
The rendered datasets are sampled with equal probability.

\begin{table}[htbp]
\centering
\small
\tabcolsep=0.5em
\caption{\textbf{Training data.} Training scenes and images of each dataset, and the labels it provides: albedo $a$, irradiance $e$, normal $n$, roughness $r$ and metallicity $m$. \emph{Geometry} marks the datasets whose cameras and metric depth we use for ground-truth tracks and to relate the exposures of a set of views. $^\ast$Exposed per image, \ie known up to one scale per image. $^\dagger$For the scenes rendered with a pinhole camera.}
\begin{tabular}{@{}l rr c ccccc c@{}}
\toprule
\textbf{Dataset} & \textbf{Scenes} & \textbf{Images} & \textbf{Resolution} & $a$ & $e$ & $n$ & $r$ & $m$ & \textbf{Geometry} \\
\midrule
Hypersim~\cite{roberts2021hypersim} & 411 & 67k & $1024\times768$ & \ding{51} & \ding{51}$^\ast$ & \ding{51} & -- & -- & \ding{51}$^\dagger$ \\
InteriorVerse~\cite{zhu2022learning} & 3{,}966 & 100k & $640\times480$ & \ding{51} & -- & \ding{51} & \ding{51} & \ding{51} & -- \\
In-house A & 125 & 162k & $1820\times1024$ & \ding{51} & \ding{51}$^\ast$ & \ding{51} & \ding{51} & -- & \ding{51} \\
In-house B & 761 & 617k & $1820\times1024$ & \ding{51} & -- & \ding{51} & \ding{51} & \ding{51} & -- \\
In-house C & 479 & 81k & $1920\times1080$ & \ding{51} & \ding{51} & \ding{51} & \ding{51} & \ding{51} & \ding{51} \\
\midrule
Photographs (pseudo-labels) & -- & 1.06M & 768 (short side) & \ding{51} & \ding{51}$^\ast$ & \ding{51} & -- & -- & -- \\
\bottomrule
\end{tabular}
\label{tab:app:data}
\end{table}

\section{Evaluation Details}
\label{sec:app:eval}

\paragraph{Test sets.}
On Hypersim~\cite{roberts2021hypersim}, we use the 46 scenes of the official test split with up to two camera trajectories each, and form 69 sets of eight views: a random frame of a trajectory and the seven frames closest to it in camera pose.
On InteriorVerse~\cite{zhu2022learning}, we form sets of up to eight views in the same way, one for each of 105 random scenes of the official test split (773 images).
On MAW~\cite{wu2023measured}, we use all 874 photographs of its 38 scenes, in sets of up to eight in capture order.
On ETH3D~\cite{schops2017multi}, we use the 13 scenes with laser scans of the high-resolution multi-view benchmark, in 50 sets of up to eight overlapping views (395 images).
The photographs of MAW and ETH3D are downsampled to a width of 1024\,px.

\paragraph{Baselines.}
We run every baseline with its released code and weights and follow its conventions for the color space of its inputs and outputs.
The video models~\cite{liang2025diffusionrenderer, vrgbx2025, sun2026unid} receive the views of a set as the frames of one clip; for the single-image results of~\cref{tab:perview}, every method processes each image separately.
MVID~\cite{du2026mvid} predicts the normals of a set in the camera frame of its first view.
We rotate them into the frame of each view with the ground-truth cameras.
All predictions are resized to the resolution of the ground truth before they are scored.

\paragraph{Accuracy.}
Albedo errors are computed in linear RGB, and LPIPS on gamma-encoded images.
For PSNR$_\text{s}$, we fit one scale per color channel to the ground truth by least squares; the irradiance is then compared after gamma encoding.
All metrics are computed over the pixels with a valid label.

\paragraph{Disagreement.}
For two views $u\neq v$ of a set, let $\mathcal{C}_{uv}$ be the pixels $q$ of view $u$ whose surface point is also visible in view $v$, at the location $x_v(q)$, and let $M_v$ be the map that a method predicts for view $v$.
We define
\begin{equation}
  \text{RMSE}_\text{r} = \left(\frac{\sum_{u\neq v}\sum_{q\in\mathcal{C}_{uv}} \lVert M_u[q] - M_v[x_v(q)]\rVert^2}{\sum_{u\neq v}\sum_{q\in\mathcal{C}_{uv}} \lVert M_u[q] - M_v[x_v(\pi_{uv}(q))]\rVert^2}\right)^{1/2},
  \label{eq:app:rmser}
\end{equation}
where $\pi_{uv}$ is a random permutation of $\mathcal{C}_{uv}$, so that the denominator pairs every pixel with the correspondence of another pixel.
It is computed per set and averaged over the sets.
In~\cref{sec:consistency}, $\mathcal{C}_{uv}$ holds the pixels whose projection with the ground-truth depth matches the depth map of view $v$ within $5\%$; in~\cref{sec:surface}, the projections of the surface points that both views see.
For the irradiance, $M_v$ is first scaled to the exposure of $M_u$ by the median ratio of the two maps over $\mathcal{C}_{uv}$, and both maps are gamma-encoded.

\paragraph{Surface points.}
We complete the protocol of~\cref{sec:surface}.
From the unprojected depth maps of a set, we keep one point per voxel, with voxels the size of a pixel at the median depth, and draw up to $500$k of the points that at least two views see.
A view sees a point if $c_v>1/2$ in~\cref{eq:confidence}, \ie, if $\delta_v<\tau$.
All maps are read at the projections of a point by bilinear interpolation, the albedo in linear RGB and the normals in the world frame.
Irradiance maps are first brought to a common exposure per set as in~\cref{sec:app:training}, separately for the ground truth and for each method.
\method predicts the normal of a point in the mean-frame, which we map to the world frame by aligning the cameras of VGGT-$\Omega$ with the ground-truth cameras.
On Hypersim, we use the 41 sets rendered with a pinhole camera, as the others cannot be reprojected.

\paragraph{Registration of InteriorVerse.}
InteriorVerse provides depth maps and intrinsics but no camera poses.
We recover the poses of a set by point-to-plane ICP~\cite{chen1992object} on the ground-truth depth maps, optimizing all views jointly and starting from the cameras of VGGT-$\Omega$.
With the initial cameras of VGGT-$\Omega$, $41.9\%$ of the pixels of a view that are visible in another view match its depth map within $1\%$; after registration, $96.1\%$ do.
For reference, the ground-truth cameras of Hypersim reach $96.9\%$.

\section{Additional Results}
\label{sec:app:results}

\subsection{Intrinsics of Surface Points}
\label{sec:app:surface_app}

\begin{figure}[htbp]
    \centering
    \includegraphics{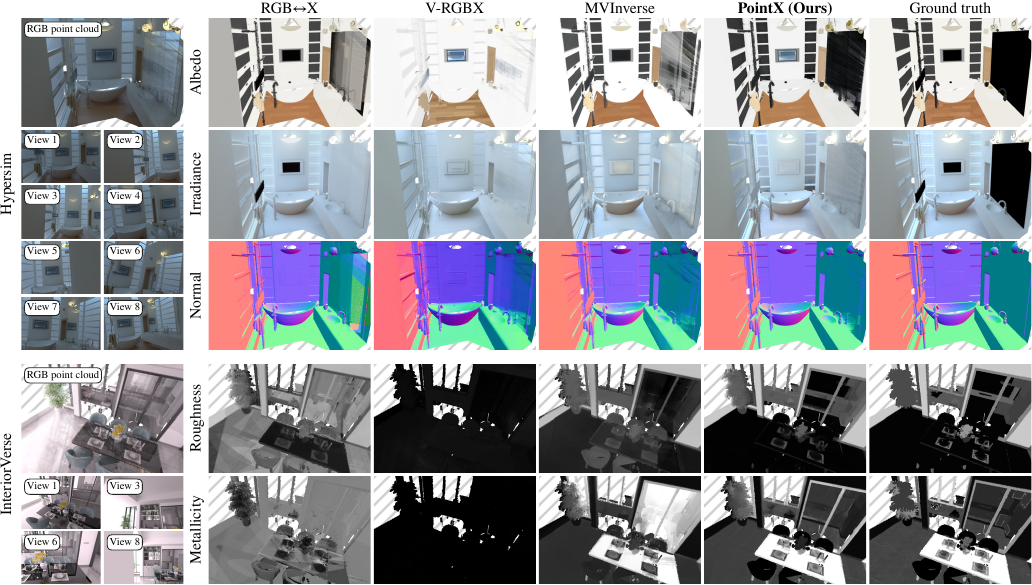}
    \caption{\textbf{Intrinsics of surface points: additional scenes.} A Hypersim~\cite{roberts2021hypersim} (top) and an InteriorVerse~\cite{zhu2022learning} (bottom) scene, shown as in~\cref{fig:qual_surface}.
    Where the views of a baseline disagree, their average shows seams and patches, \eg, on the glass panel (top) and on the floor (bottom).}
    \label{fig:app:surface1}
    \vspace{1mm}

    \includegraphics{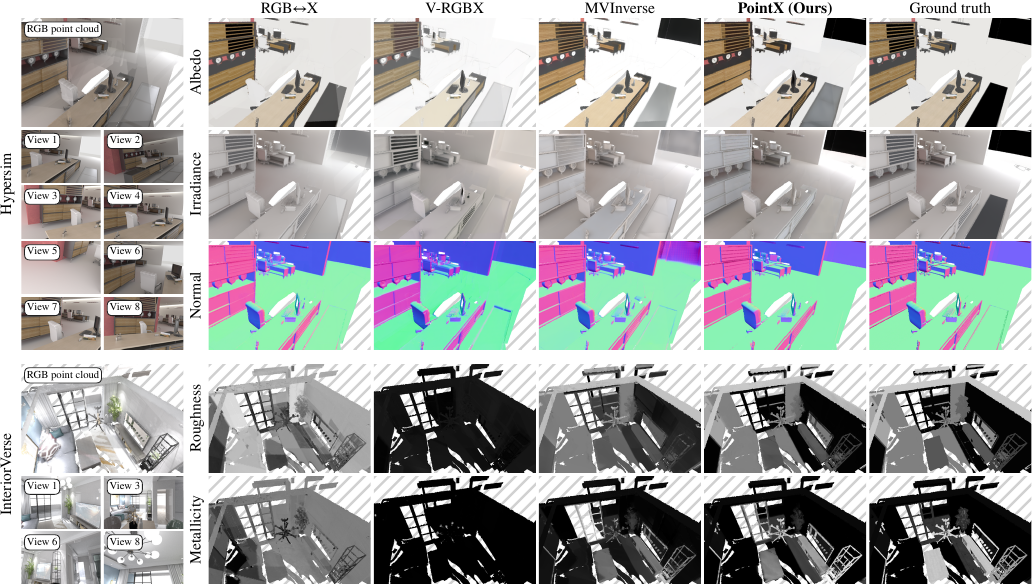}
    \caption{\textbf{Intrinsics of surface points: additional scenes (continued).} Two further scenes, shown as in~\cref{fig:app:surface1}.}
    \label{fig:app:surface2}
\end{figure}

\begin{figure}[htbp]
    \centering
    \includegraphics{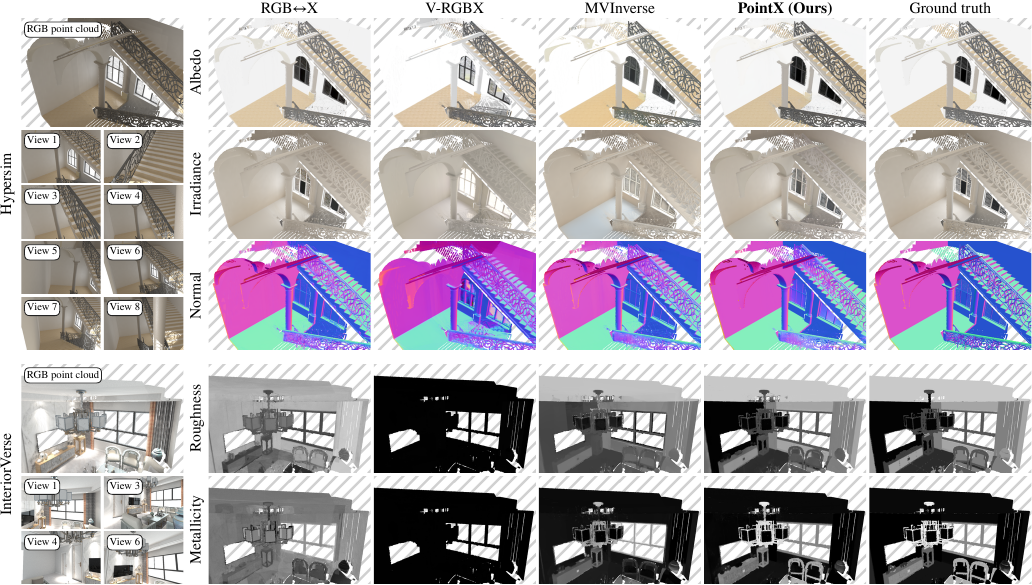}
    \caption{\textbf{Intrinsics of surface points: additional scenes (continued).} Two further scenes, shown as in~\cref{fig:app:surface1}.}
    \label{fig:app:surface3}

    \includegraphics{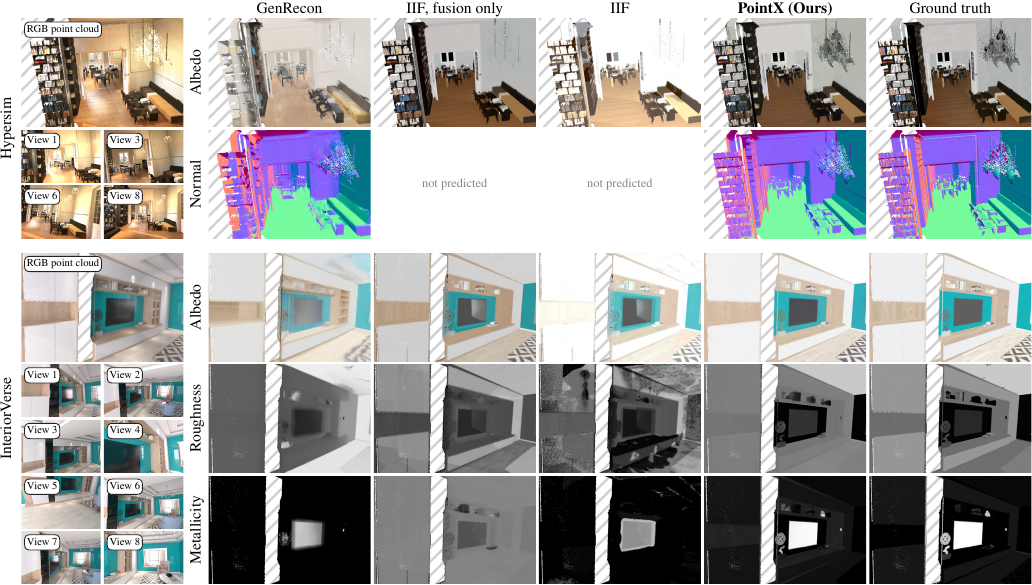}
    \caption{\textbf{Additional baselines.} A Hypersim~\cite{roberts2021hypersim} (top) and an InteriorVerse~\cite{zhu2022learning} (bottom) scene, shown as in~\cref{fig:qual_surface}, for GenRecon~\cite{schmid2026genrecon} and Intrinsic Image Fusion (IIF)~\cite{kocsis2026intrinsic} with and without its inverse path tracing.
    The materials of GenRecon look plausible but differ from the true ones.
    The inverse path tracing of IIF brightens the albedo, since much of the light in the scene comes from surfaces that the eight views do not show.}
    \label{fig:app:posed}
\end{figure}

In~\cref{sec:surface}, the per-view predictions of a baseline are combined by their unweighted mean.
\Cref{tab:app:fusion} compares this rule with a mean weighted by the confidence of~\cref{eq:confidence} and with a median.
The rule has little effect on any method.
\Cref{fig:app:surface1,fig:app:surface2,fig:app:surface3} show the intrinsic point clouds for further scenes of Hypersim~\cite{roberts2021hypersim} and InteriorVerse~\cite{zhu2022learning}.

\begin{table}[htbp]
\centering
\small
\tabcolsep=0.45em
\caption{\textbf{Fusion of per-view predictions at surface points.} Accuracy on the task of~\cref{sec:surface} when the per-view predictions of a method are combined by their mean, by their mean weighted with the confidence of~\cref{eq:confidence}, or by their median. \method predicts every point once and needs no fusion, so its value is the same under each rule. The other metrics of~\cref{tab:surface} behave alike. Best, second and third best are highlighted. $^\dagger$Trained on the Hypersim test scenes, not ranked.}
\resizebox{\linewidth}{!}{
\begin{tabular}{@{}l ccc ccc ccc ccc@{}}
\toprule
\multirow{3}[4]{*}{\textbf{Method}} & \multicolumn{6}{c}{\textbf{Hypersim}~\cite{roberts2021hypersim}} & \multicolumn{6}{c}{\textbf{InteriorVerse}~\cite{zhu2022learning}} \\
\cmidrule(lr){2-7}\cmidrule(lr){8-13}
 & \multicolumn{3}{c}{\textbf{Albedo} PSNR~$\uparrow$} & \multicolumn{3}{c}{\textbf{Normal} Mean$^\circ$~$\downarrow$} & \multicolumn{3}{c}{\textbf{Albedo} PSNR~$\uparrow$} & \multicolumn{3}{c}{\textbf{Normal} Mean$^\circ$~$\downarrow$} \\
\cmidrule(lr){2-4}\cmidrule(lr){5-7}\cmidrule(lr){8-10}\cmidrule(lr){11-13}
 & Mean & Weighted & Median & Mean & Weighted & Median & Mean & Weighted & Median & Mean & Weighted & Median \\
\midrule
Intrinsic Image Diffusion~\cite{kocsis2024intrinsic} & 12.37 & 12.37 & 12.36 & \multicolumn{3}{c}{--} & 16.11 & 16.12 & 16.10 & \multicolumn{3}{c}{--} \\
RGB$\leftrightarrow$X$^\dagger$~\cite{zeng2024rgb} & \textcolor{gray}{24.09} & \textcolor{gray}{24.09} & \textcolor{gray}{24.33} & \textcolor{gray}{8.0} & \textcolor{gray}{8.0} & \textcolor{gray}{7.8} & 9.57 & 9.57 & 9.55 & 12.3 & 12.2 & 11.4 \\
DiffusionRenderer~\cite{liang2025diffusionrenderer} & 14.50 & 14.49 & 14.49 & 13.6 & 13.5 & 13.0 & 10.82 & 10.82 & 10.83 & 11.4 & 11.3 & 10.7 \\
V-RGBX~\cite{vrgbx2025} & 11.98 & 11.99 & 11.93 & 23.3 & 23.3 & 23.3 & 11.42 & 11.42 & 11.34 & 22.4 & 22.4 & 22.2 \\
MVInverse~\cite{wu2026mvinverse} & 18.62 & 18.61 & 18.62 & \best{8.2} & \best{8.2} & \best{7.9} & 18.41 & 18.43 & 18.41 & \secondbest{7.8} & \secondbest{7.7} & \secondbest{7.5} \\
RGB$\leftrightarrow$X + GeoID$^\dagger$~\cite{dirik2026geoid} & \textcolor{gray}{23.62} & \textcolor{gray}{23.63} & \textcolor{gray}{23.82} & \multicolumn{3}{c}{--} & 10.89 & 10.89 & 10.87 & \multicolumn{3}{c}{--} \\
UniD~\cite{sun2026unid} & 17.34 & 17.34 & 17.33 & 14.9 & 14.8 & 14.6 & 9.63 & 9.63 & 9.65 & 15.7 & 15.7 & 15.3 \\
Marigold V2~\cite{pavlovic2026marigoldv2} & \secondbest{21.16} & \secondbest{21.17} & \secondbest{21.31} & \thirdbest{8.4} & \thirdbest{8.4} & \secondbest{8.2} & 11.57 & 11.58 & 11.60 & \thirdbest{8.7} & \thirdbest{8.6} & \thirdbest{8.4} \\
MVID~\cite{du2026mvid} & \best{26.22} & \best{26.24} & \best{26.38} & 8.9 & 8.8 & 8.6 & \secondbest{24.64} & \secondbest{24.67} & \secondbest{24.67} & 24.3 & 24.3 & 24.3 \\
MVID-$\Omega$~\cite{du2026mvid} & 17.72 & 17.73 & 17.73 & 11.5 & 11.5 & 11.3 & \thirdbest{19.32} & \thirdbest{19.33} & \thirdbest{19.31} & 25.4 & 25.4 & 25.3 \\
\textit{\method w/o pooling} & 20.33 & 20.33 & 20.34 & 8.6 & 8.6 & 8.4 & 25.41 & 25.44 & 25.47 & 5.7 & 5.6 & 5.4 \\
\midrule
\textbf{\method (Ours)} & \thirdbest{20.32} & \thirdbest{20.32} & \thirdbest{20.32} & \secondbest{8.3} & \secondbest{8.3} & \thirdbest{8.3} & \best{25.50} & \best{25.50} & \best{25.50} & \best{5.2} & \best{5.2} & \best{5.2} \\
\bottomrule
\end{tabular}
}
\label{tab:app:fusion}
\end{table}

\clearpage

\subsection{Additional Baselines}
\label{sec:app:baselines_app}

In~\cref{sec:experiments}, we omit methods that need posed images or per-scene optimization.
Since the task of~\cref{sec:surface} provides the cameras and the geometry of every scene, we evaluate two of them on it: Intrinsic Image Fusion~\cite{kocsis2026intrinsic} and GenRecon~\cite{schmid2026genrecon}.
\Cref{tab:app:posed} shows that \method is considerably more accurate than both on every property of InteriorVerse.
On Hypersim, it is more accurate than GenRecon, while Intrinsic Image Fusion builds on RGB$\leftrightarrow$X~\cite{zeng2024rgb}, which was trained on the test scenes.
\Cref{fig:app:posed} shows two scenes.

\paragraph{Setup.}
Intrinsic Image Fusion draws samples of RGB$\leftrightarrow$X for every view, fuses them into one material field on a mesh of the scene, and refines this field by inverse path tracing.
We give it the ground-truth cameras, the ground-truth depth maps as its mesh, and the instance segmentation it requires, from SAM~2~\cite{ravi2025sam} masks that we link across views with the ground-truth depth.
GenRecon generates a mesh with material textures from posed images and sparse scene points, for which we use the ground-truth cameras and points of the ground-truth depth maps.
The result of each method is rendered into the views of a set and scored like the maps of the other baselines.
Intrinsic Image Fusion optimizes a set for more than one hour on an A100 GPU and GenRecon takes about ten minutes, whereas \method takes seconds (\cref{sec:ablations}).

\paragraph{Intrinsic Image Fusion.}
The fusion makes the samples of RGB$\leftrightarrow$X about as consistent as the ground-truth maps (\cref{tab:surface}), but it does not correct them: its accuracy remains that of RGB$\leftrightarrow$X.
The inverse path tracing explains the images by the light that the surfaces of the mesh exchange.
Its authors evaluate it on scenes covered by about 200 views, whereas our sets hold eight views of a part of a scene, so much of this light comes from surfaces that no view shows.
The optimization compensates for the missing light with a brighter albedo, which saturates on Hypersim.
On InteriorVerse, it improves the metallicity but degrades the roughness and the PSNR$_\text{s}$ of the albedo.
On one of the scenes of its authors, its released pipeline raises the albedo PSNR of the fusion from 15.0 to 22.6\,dB with all 202 views, and not at all with eight of them.

\paragraph{GenRecon.}
GenRecon is a generative reconstruction model, and its authors evaluate its geometry only.
Its materials are sampled from a 3D generative prior conditioned on the images: they look plausible but are far from the true ones.
Its mesh also deviates from the true surface, in particular on Hypersim, where a set often shows only a part of a large room, so the renders of this one mesh still disagree at the true surface points.

\begin{table}[htbp]
\centering
\small
\tabcolsep=0.45em
\caption{\textbf{Additional baselines.} The protocol of~\cref{tab:surface} for two methods that require posed images, GenRecon~\cite{schmid2026genrecon} and Intrinsic Image Fusion~\cite{kocsis2026intrinsic}. Intrinsic Image Fusion fuses samples of RGB$\leftrightarrow$X, listed for reference, and refines the result by inverse path tracing; \emph{fusion only} omits this refinement. Neither method predicts irradiance, and Intrinsic Image Fusion predicts no normals. The best result is highlighted. $^\dagger$Based on RGB$\leftrightarrow$X, which was trained on the Hypersim test scenes; not ranked.}
\resizebox{\linewidth}{!}{
\begin{tabular}{@{}l ccc GG ccccc GGGG@{}}
\toprule
\multirow{4}[6]{*}{\textbf{Method}} & \multicolumn{5}{c}{\textbf{Hypersim}~\cite{roberts2021hypersim}} & \multicolumn{9}{c}{\textbf{InteriorVerse}~\cite{zhu2022learning}} \\
\cmidrule(lr){2-6}\cmidrule(lr){7-15}
 & \multicolumn{3}{c}{\textbf{Accuracy}} & \multicolumn{2}{G}{\textbf{Disagreement}} & \multicolumn{5}{c}{\textbf{Accuracy}} & \multicolumn{4}{G}{\textbf{Disagreement}} \\
\cmidrule(lr){2-4}\cmidrule(lr){5-6}\cmidrule(lr){7-11}\cmidrule(lr){12-15}
 & \multicolumn{2}{c}{\textbf{Albedo}} & \textbf{Normal} & \textbf{Albedo} & \textbf{Normal} & \multicolumn{2}{c}{\textbf{Albedo}} & \textbf{Normal} & \textbf{Rough.} & \textbf{Metal.} & \textbf{Albedo} & \textbf{Normal} & \textbf{Rough.} & \textbf{Metal.} \\
\cmidrule(lr){2-3}\cmidrule(lr){4-4}\cmidrule(lr){5-5}\cmidrule(lr){6-6}\cmidrule(lr){7-8}\cmidrule(lr){9-9}\cmidrule(lr){10-10}\cmidrule(lr){11-11}\cmidrule(lr){12-12}\cmidrule(lr){13-13}\cmidrule(lr){14-14}\cmidrule(lr){15-15}
 & PSNR~$\uparrow$ & PSNR$_\text{s}$~$\uparrow$ & Mean$^\circ$~$\downarrow$ & RMSE$_\text{r}$~$\downarrow$ & Mean$^\circ$~$\downarrow$ & PSNR~$\uparrow$ & PSNR$_\text{s}$~$\uparrow$ & Mean$^\circ$~$\downarrow$ & RMSE~$\downarrow$ & RMSE~$\downarrow$ & RMSE$_\text{r}$~$\downarrow$ & Mean$^\circ$~$\downarrow$ & RMSE$_\text{r}$~$\downarrow$ & RMSE$_\text{r}$~$\downarrow$ \\
\midrule
GenRecon~\cite{schmid2026genrecon} & 11.85 & 13.61 & 23.4 & 0.472 & 13.9 & 14.66 & 15.30 & 12.5 & 0.446 & 0.186 & 0.360 & \best{9.8} & 0.299 & 0.487 \\
RGB$\leftrightarrow$X$^\dagger$~\cite{zeng2024rgb} & \textcolor{gray}{24.09} & \textcolor{gray}{24.89} & \textcolor{gray}{8.0} & \textcolor{gray}{0.196} & \textcolor{gray}{8.3} & 9.57 & 18.96 & 12.3 & 0.304 & 0.407 & 0.499 & 18.1 & 0.582 & 0.837 \\
Intrinsic Image Fusion$^\dagger$~\cite{kocsis2026intrinsic}, fusion only & \textcolor{gray}{24.63} & \textcolor{gray}{25.52} & -- & \textcolor{gray}{0.068} & -- & 10.61 & 19.10 & -- & 0.281 & 0.432 & \best{0.106} & -- & \best{0.096} & 0.191 \\
Intrinsic Image Fusion$^\dagger$~\cite{kocsis2026intrinsic} & \textcolor{gray}{8.18} & \textcolor{gray}{15.26} & -- & \textcolor{gray}{0.097} & -- & 13.78 & 16.73 & -- & 0.359 & 0.204 & 0.124 & -- & 0.107 & \best{0.137} \\
\midrule
\textbf{\method (Ours)} & \best{20.32} & \best{22.76} & \best{8.3} & \multicolumn{2}{c}{\cellcolor{constructgreen}\textbf{0} by construction} & \best{25.50} & \best{25.97} & \best{5.2} & \best{0.083} & \best{0.072} & \multicolumn{4}{c}{\cellcolor{constructgreen}\textbf{0} by construction} \\
\bottomrule
\end{tabular}
}
\label{tab:app:posed}
\end{table}

\clearpage

\subsection{Consistency of Per-View Maps}
\label{sec:app:consistency_app}

\begin{figure}[htbp]
    \centering
    \includegraphics{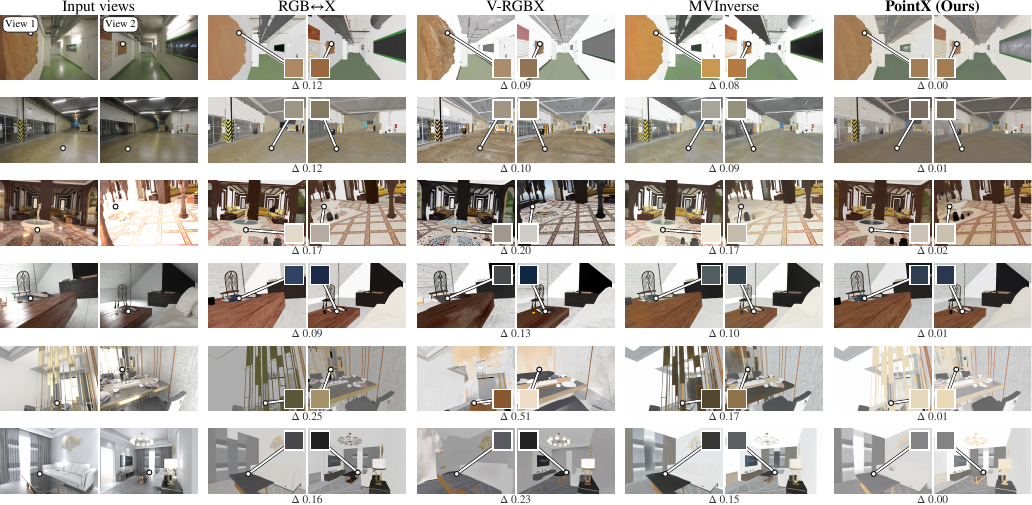}
    \caption{\textbf{Consistency of per-view maps: additional scenes.} Albedo predictions for two views of a scene on ETH3D~\cite{schops2017multi} (rows 1--2), Hypersim~\cite{roberts2021hypersim} (rows 3--4) and InteriorVerse~\cite{zhu2022learning} (rows 5--6), shown as in~\cref{fig:qual_consistency} (\method: RoMa-$\Omega$ tracks). The marked surface point is the same in both views; the insets show the value each method predicts for it, and $\Delta$ their mean absolute difference.}
    \label{fig:app:consistency}
\end{figure}

\Cref{tab:app:consistency} applies the protocol of~\cref{sec:consistency} to the sets of Hypersim and InteriorVerse used in~\cref{sec:surface}, with their ground-truth depth and, for InteriorVerse, the registered cameras.
The results agree with those on ETH3D.
\Cref{fig:app:consistency} compares the albedo that each method predicts for the same surface point in two views, on all three datasets.

\begin{table}[htbp]
\centering
\small
\tabcolsep=0.5em
\caption{\textbf{Consistency of per-view maps on rendered scenes.} The protocol of~\cref{tab:consistency} on Hypersim and InteriorVerse: disagreement between the maps of two views at ground-truth correspondences. Gray: ground-truth geometry at test time, or $^\dagger$trained on the Hypersim test scenes; not ranked. Best, second and third best are highlighted.}
\resizebox{\linewidth}{!}{
\begin{tabular}{@{}l c ccccc ccccc@{}}
\toprule
\multirow{3}[4]{*}{\textbf{Method}} & \multirow{3}[4]{*}{\textbf{Track Source}} & \multicolumn{5}{c}{\textbf{Hypersim}~\cite{roberts2021hypersim}} & \multicolumn{5}{c}{\textbf{InteriorVerse}~\cite{zhu2022learning}} \\
\cmidrule(lr){3-7}\cmidrule(lr){8-12}
 & & \textbf{Albedo} & \textbf{Irrad.} & \textbf{Normal} & \textbf{Rough.} & \textbf{Metal.} & \textbf{Albedo} & \textbf{Irrad.} & \textbf{Normal} & \textbf{Rough.} & \textbf{Metal.} \\
\cmidrule(lr){3-3}\cmidrule(lr){4-4}\cmidrule(lr){5-5}\cmidrule(lr){6-6}\cmidrule(lr){7-7}\cmidrule(lr){8-8}\cmidrule(lr){9-9}\cmidrule(lr){10-10}\cmidrule(lr){11-11}\cmidrule(lr){12-12}
 & & RMSE$_\text{r}$~$\downarrow$ & RMSE$_\text{r}$~$\downarrow$ & Mean$^\circ$~$\downarrow$ & RMSE$_\text{r}$~$\downarrow$ & RMSE$_\text{r}$~$\downarrow$ & RMSE$_\text{r}$~$\downarrow$ & RMSE$_\text{r}$~$\downarrow$ & Mean$^\circ$~$\downarrow$ & RMSE$_\text{r}$~$\downarrow$ & RMSE$_\text{r}$~$\downarrow$ \\
\midrule
\multicolumn{12}{@{}l}{\textit{Single-view methods}} \\
Intrinsic Image Diffusion~\cite{kocsis2024intrinsic} & -- & 0.469 & -- & -- & 0.641 & 0.596 & 0.453 & -- & -- & 0.582 & 0.512 \\
RGB$\leftrightarrow$X$^\dagger$~\cite{zeng2024rgb} & -- & \textcolor{gray}{0.228} & \textcolor{gray}{0.321} & \textcolor{gray}{10.8} & \textcolor{gray}{0.409} & \textcolor{gray}{0.643} & 0.515 & 0.434 & 21.8 & 0.605 & 0.838 \\
Marigold V2~\cite{pavlovic2026marigoldv2} & -- & 0.238 & -- & 7.6 & -- & -- & 0.403 & -- & 11.6 & -- & -- \\
\midrule
\multicolumn{12}{@{}l}{\textit{Video \& Multi-view methods}} \\
DiffusionRenderer~\cite{liang2025diffusionrenderer} & -- & 0.326 & -- & 13.5 & 0.514 & 0.892 & 0.520 & -- & 15.2 & 0.813 & 0.948 \\
V-RGBX~\cite{vrgbx2025} & -- & 0.264 & 0.324 & 13.4 & 0.484 & 0.906 & 0.471 & 0.386 & 18.5 & 0.805 & 0.961 \\
MVInverse~\cite{wu2026mvinverse} & -- & \secondbest{0.174} & 0.326 & 8.0 & \secondbest{0.244} & \best{0.244} & 0.249 & 0.340 & \thirdbest{10.1} & \thirdbest{0.250} & \thirdbest{0.349} \\
RGB$\leftrightarrow$X + GeoID$^\dagger$~\cite{dirik2026geoid} & -- & \textcolor{gray}{0.221} & -- & -- & \textcolor{gray}{0.368} & \textcolor{gray}{0.622} & 0.505 & -- & -- & 0.573 & 0.823 \\
UniD~\cite{sun2026unid} & -- & 0.221 & 0.303 & 12.2 & -- & -- & 0.364 & 0.330 & 19.2 & -- & -- \\
MVID~\cite{du2026mvid} & -- & \secondbest{0.174} & \thirdbest{0.261} & 6.7 & -- & -- & \thirdbest{0.196} & 0.285 & 14.1 & -- & -- \\
MVID-$\Omega$~\cite{du2026mvid} & -- & 0.201 & 0.277 & \thirdbest{6.6} & -- & -- & 0.217 & \thirdbest{0.278} & 10.2 & -- & -- \\
\multirow{3}{*}{\textbf{\method (Ours)}} & VGGT-$\Omega$~\cite{wang2026vggtomega} & 0.195 & \secondbest{0.198} & \secondbest{5.7} & \thirdbest{0.246} & \thirdbest{0.317} & \secondbest{0.195} & \secondbest{0.214} & \secondbest{4.9} & \secondbest{0.145} & \secondbest{0.285} \\
 & RoMa-$\Omega$~\cite{nordstrom2026romaomega} & \best{0.135} & \best{0.155} & \best{4.0} & \best{0.190} & \secondbest{0.257} & \best{0.161} & \best{0.200} & \best{4.3} & \best{0.126} & \best{0.233} \\
 & Ground truth & \textcolor{gray}{0.076} & \textcolor{gray}{0.080} & \textcolor{gray}{1.5} & \textcolor{gray}{0.083} & \textcolor{gray}{0.118} & \textcolor{gray}{0.110} & \textcolor{gray}{0.132} & \textcolor{gray}{2.1} & \textcolor{gray}{0.075} & \textcolor{gray}{0.155} \\
\bottomrule
\end{tabular}
}
\label{tab:app:consistency}
\end{table}

\subsection{Accuracy of Per-View Maps}
\label{sec:app:accuracy_app}

\begin{figure}[htbp]
\centering
\includegraphics[width=0.5\linewidth]{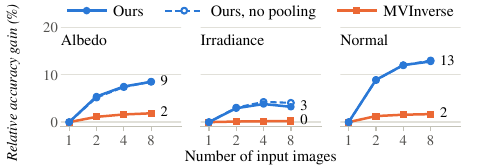}
\caption{\textbf{Accuracy with more input images on Hypersim~\cite{roberts2021hypersim}.} Reduction of the error relative to a single image, as in~\cref{fig:views} (RMSE; mean angle for normals). \method uses RoMa-$\Omega$ tracks.}
\label{fig:app:views}
\end{figure}

\Cref{fig:app:views} repeats the experiment of~\cref{fig:views} on Hypersim.
The views of a set are split into consecutive groups of $N$ views that are predicted independently, so that every view is predicted from $N$ images and the same views are scored for every $N$.
As on InteriorVerse, every property predicted by \method becomes more accurate with more images, MVInverse gains considerably less, and the gain remains without pooling.

\subsection{Ablation Studies}
\label{sec:app:ablations_app}

\begin{figure}[htbp]
    \centering
    \includegraphics{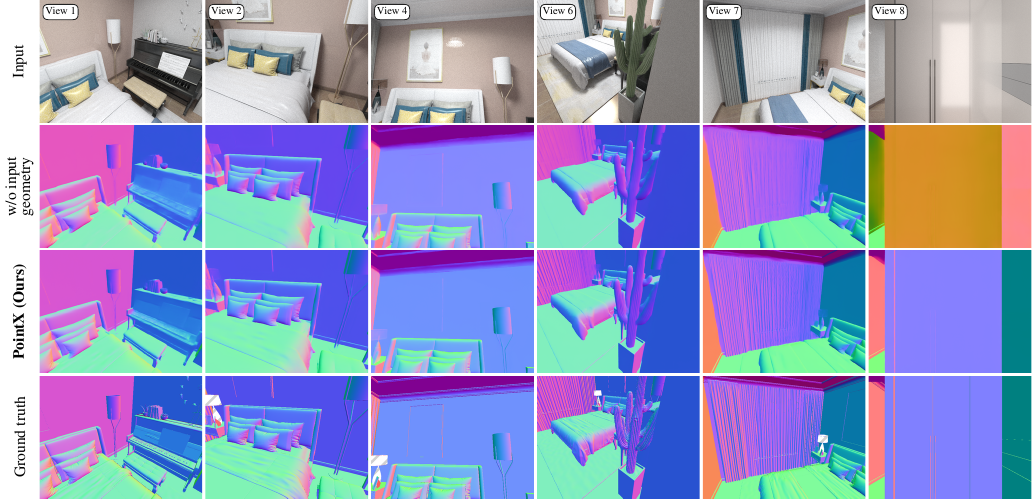}
    \caption{\textbf{Normals without the geometry channels.} Six of the eight views of an InteriorVerse~\cite{zhu2022learning} set and their normal maps, predicted from all eight images (RoMa-$\Omega$ tracks). Without the geometry channels, the normals of view 8, which shares little content with the other views, are predicted in a wrong orientation. Regions without ground truth are hatched.}
    \label{fig:app:normals}
\end{figure}

Without the geometry channels, the model must infer the orientation of each camera in the mean-frame from the images alone (\cref{sec:ablations}).
\Cref{fig:app:normals} shows a set where this fails: view 8 shares little content with the other views, and its normals are predicted in a wrong orientation.
With the geometry channels, the normals of all views are oriented correctly.
In most sets the effect is smaller, a slight tilt of the normals of individual views.

\clearpage

\section{Relighting and Editing}
\label{sec:app:applications}

The point cloud of~\cref{fig:teaser} holds one \method query per point of the laser scan.

\paragraph{Relighting.}
Let $L$ be the product of the predicted albedo and irradiance of a point, \ie, its diffuse radiance under the original illumination.
We dim it and add the radiance $L_\text{new}$ of the point under new point lights, computed with a Cook--Torrance model~\cite{cook1982reflectance} from its predicted albedo, roughness, metallicity and normal: $L' = kL + L_\text{new}$, with $k<1$.
Each photograph is then multiplied, in linear RGB, by $L'/L$ at the pixels to which the points project.

\paragraph{Albedo editing.}
We scale the albedo of the points of an object and multiply each photograph by the same factor, in linear RGB, at the pixels to which these points project.

\section{Limitations and Future Work}
\label{sec:app:limitations}

\paragraph{Geometry and tracks.}
\method locates a surface point through the geometry or the matches it is given, so their errors carry over to its predictions.
With the geometry estimated by VGGT-$\Omega$~\cite{wang2026vggtomega}, its per-view maps are less consistent than with the tracks of RoMa-$\Omega$~\cite{nordstrom2026romaomega} or with ground-truth geometry (\cref{tab:consistency}), and an error in an estimated camera tilts the normals of its view.
RoMa-$\Omega$ in turn matches every pair of views, so its cost grows quadratically with their number.
Since the tracks are an input, \method benefits from any progress in reconstruction and matching without retraining.

\paragraph{Pooling.}
The weights of~\cref{eq:pool} have no learned parameters and depend only on the visibility of a point.
They make the prediction of a point independent of the view, but not smooth along a surface: neighboring points that are visible in different sets of views are decoded from different sets of tokens, which can leave a small step between their predictions, \eg, where the frame of another image ends.
A confidence that also decreases towards the border of an image is a simple extension that we leave for future work.

\paragraph{Albedo and illumination.}
On Hypersim, \method does not reach the most accurate baseline on albedo and irradiance (\cref{tab:surface,tab:perview}).
Part of this gap is the overall scale of the albedo, \ie, the difference between PSNR and PSNR$_\text{s}$, and it remains when more views are added.
Views of a scene under the same illumination add evidence about its surfaces, but not about how their brightness divides into albedo and irradiance.
\method also assumes a static scene whose illumination does not change between the images.
Images captured under varying illumination could provide the missing evidence and are an interesting direction.

\paragraph{Training data.}
All multi-view supervision comes from rendered scenes: the real photographs are single images with pseudo-labels, and they carry no roughness or metallicity (\cref{tab:app:data}).
On real scenes, we evaluate the accuracy of the albedo and the consistency of all five properties, but not the accuracy of the other properties.
Real multi-view captures with measured materials would benefit both training and evaluation.

\paragraph{Number of views.}
\method encodes all images of a scene in one forward pass and is trained with up to 24 views.
Its consistency holds by construction within such a set of images. Extending the pooling across sets, for long videos or large scenes, is left for future work.

\end{document}